\documentclass[11pt]{article}
\usepackage{acl}
\usepackage{times}
\usepackage{latexsym}
\usepackage{booktabs}
\usepackage{multirow}
\usepackage{graphicx}
\usepackage{amsmath}
\usepackage{url}
\usepackage{xcolor}
\usepackage{array}
\usepackage{tabularx}
\usepackage{adjustbox}
\usepackage{makecell}
\usepackage{capt-of}
\graphicspath{{figures/}}
\title{Beyond Agreement: Scoring Panel-Surfaced Biomedical Entity Candidates for Curator Triage}

\author{
\begin{tabular}{@{}c@{}}
\textbf{Shuheng Cao}$^{1,*}$ \quad
\textbf{Ruiqi Chen}$^{2,*}$ \quad
\textbf{Renjie Cao}$^{3,\dagger}$ \\
\textbf{Zhenhao Zhang}$^{4,\dagger}$ \quad
\textbf{Siyu Zhang}$^{1,\dagger}$ \quad
\textbf{Tingting Dan}$^{5,\ddagger}$ \\[0.5em]
{\small $^{1}$University of California, San Diego} \\
{\small $^{2}$University of Michigan, Ann Arbor} \\
{\small $^{3}$The Hong Kong University of Science and Technology, Guangzhou} \\
{\small $^{4}$ShanghaiTech University} \\
{\small $^{5}$University of North Carolina, Chapel Hill} \\[0.4em]
{\small $^{*}$Equal contribution as co-first authors.} \\
{\small $^{\dagger}$Equal contribution.} \\
{\small $^{\ddagger}$Corresponding author.}
\end{tabular}
}

\begin{document}
\maketitle

\begin{abstract}
Biomedical curation turns entity mentions into review decisions. Modern LLM panels can surface plausible biomedical candidates, but multi-model agreement signals salience and does not establish corpus convention correctness. We define panel surfaced candidate verification as a curator-facing layer after candidate generation, and introduce a candidate-level panel output benchmark that aligns eight LLMs' outputs over five public biomedical Named Entity Recognition(NER) datasets into candidate rows. We instantiate this layer with BioConCal, a supervised scorer that uses inference-time gold-free panel and candidate features to rank a candidate stream. In the in-domain document split, BioConCal raises AUROC from 0.753 for raw agreement to 0.910. At a validation selected 0.95 precision target, it selects 1{,}340 candidate rows at empirical precision 0.939, compared with 293 rows for raw agreement, with $R_{\mathrm{cand}}=0.592$ and row label $R_{\mathrm{corpus}}=0.523$ against a panel row label ceiling of 0.883. These results show that calibrated panel evidence can turn a noisy candidate stream into a higher-yield review queue, while preserving the workflow boundary that candidate sources set the recoverable universe. Target domain validation, source expansion tradeoffs, span reconciliation, and human review remain downstream requirements.
\end{abstract}

\section{Introduction}
Biomedical named entity recognition is often evaluated as span extraction. In curation pipelines, however, extracted mentions become review decisions. A curator must decide whether a candidate should be accepted, routed for review, or rejected. False positives that reach a curated resource can be costly to trace and reverse \citep{li2016bc5cdr,dogan2014ncbi,smith2008biocreative,collier2004jnlpba,krallinger2015chemdner}.

Modern LLMs sharpen this decision point. They can surface plausible biomedical candidates under new entity definitions without task-specific training \citep{singhal2023medpalm,luo2022biogpt,nagar2024llms}. Their outputs also include plausible non gold candidates, boundary variants, type confusions, and convention mismatches. The central difficulty is not only whether a phrase looks biomedical. It is whether the candidate follows the target corpus convention. We use corpus convention correctness to mean compatibility with the target dataset boundary, entity type, granularity, normalization, occurrence, and annotation rules. It does not mean independent biomedical truth verification.

Multi-model agreement is an obvious triage signal. Candidates emitted by several models are often more salient. Yet salience is not correctness under a corpus convention. The question for curation is whether a stream of candidates surfaced by a model panel can be calibrated and prioritized for curator review.

We call this problem panel surfaced candidate verification. It is a layer after candidate generation and before human curation. The input is a fixed candidate stream from a defined model panel. The output is a candidate correctness score for the curator facing triage. The unit is an aligned candidate row, not an individual LLM extractor output. The panel defines the recoverable candidate universe, and scoring determines review yield within that universe. We therefore separate candidate-level triage from corpus-level coverage.

\begin{figure*}[t]
\centering
\includegraphics[width=\textwidth]{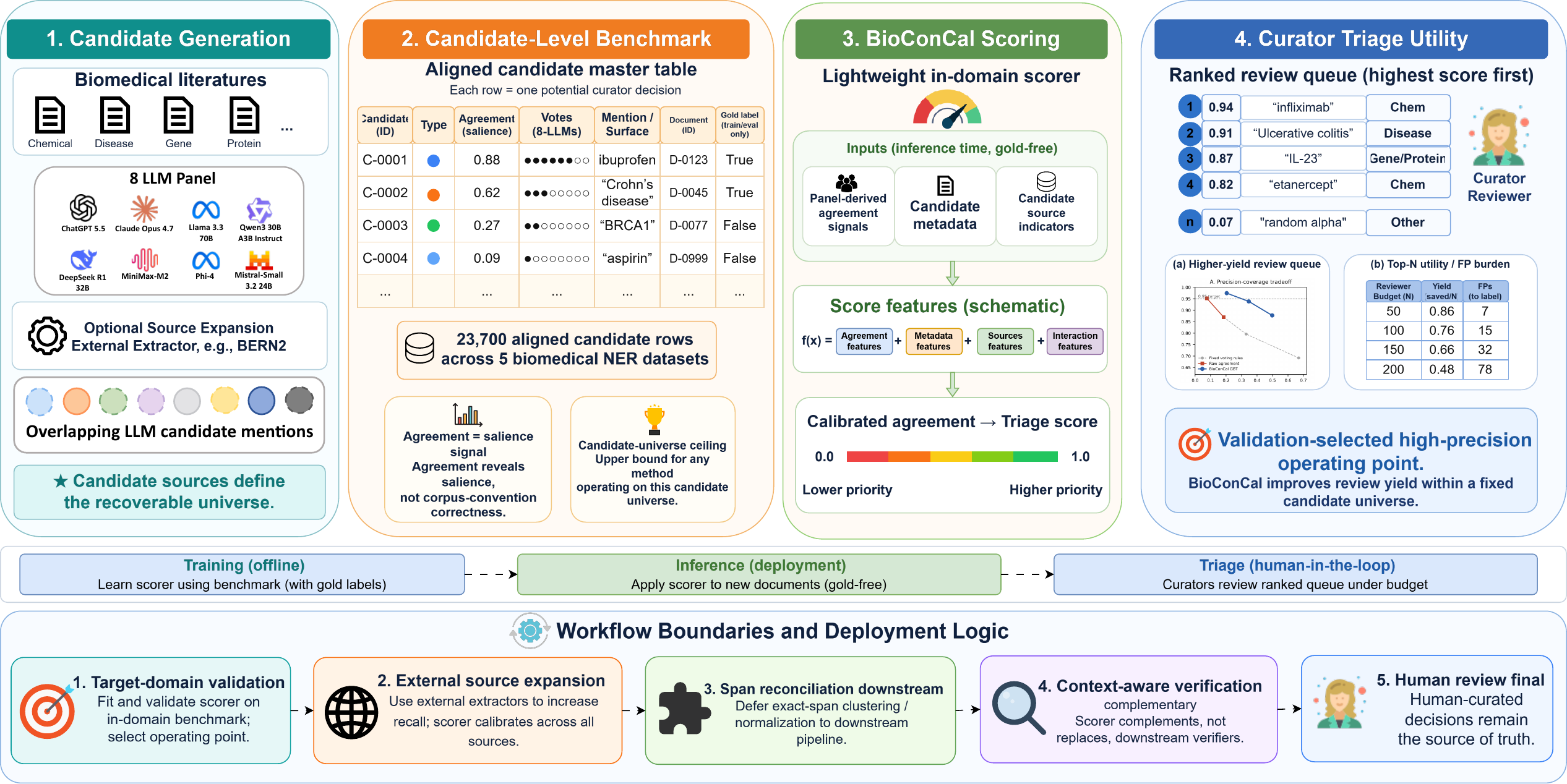}
\caption{BioConCal overview. A multi-model panel first surfaces biomedical entity candidates, which are aligned into candidate rows for benchmark construction. BioConCal scores this fixed candidate stream using inference-time gold-free panel and candidate features, then ranks candidates for curator triage. The workflow separates candidate generation, candidate-level verification, validation-selected operating points, span reconciliation, and final human review.}
\label{fig:teaser}
\end{figure*}
We instantiate this setting with a candidate-level panel output benchmark and BioConCal. The benchmark aligns eight LLMs' outputs over five public biomedical NER datasets into candidate rows with agreement structure, inference time gold-free candidate features, and multiple correctness labels. BioConCal is an in-domain supervised scorer learned from labelled candidate rows. At inference time, it uses panel-derived signals and candidate metadata to prioritize a fixed candidate stream for review. In domain, BioConCal raises AUROC from 0.753 for raw agreement to 0.910 and selects 1{,}340 candidate rows at empirical precision 0.939 under the validation selected 0.95 target, compared with 293 rows for raw agreement.

We make four contributions.
\begin{enumerate}
    \item We define panel surfaced candidate verification, a curator facing problem that starts after candidate generation and asks which surfaced candidates should be prioritized for review.
    \item We construct a candidate-level panel output benchmark by aligning eight LLMs' outputs over five public biomedical NER datasets into candidate rows with agreement structure, candidate metadata, correctness labels, and candidate universe ceilings.
    \item We introduce \textbf{BioConCal}, an in-domain supervised scorer that uses inference time gold free panel and candidate features to calibrate fixed stream candidates for curator triage.
    \item We analyze the workflow boundary, showing how agreement errors, validation-selected operating points, source expansion, deterministic span reconciliation, and human review shape deployment.
\end{enumerate}

\section{Benchmark and Candidate Construction}
This benchmark evaluates candidate verification after generation. Its
unit is an aligned candidate row surfaced by an explicitly defined panel. Each row represents a possible curator decision rather than a standalone extractor output.

\subsection{Datasets}
We use five biomedical NER datasets covering eight unified entity types (chemical, disease, gene, protein, DNA, RNA, cell line, cell type). Table~\ref{tab:datasets} lists the datasets. Four datasets form the main benchmark. CHEMDNER is treated as robustness only for offset-sensitive analyses because the public mirror does not provide reliable character offsets. For each dataset, we sample 500 documents with a fixed random seed. This fixed-size design balances entity schemas and keeps the eight model panel evaluation tractable. The benchmark is not intended to replace full corpus extractor leaderboards. It is designed to study candidate verification under a shared panel protocol. The resulting benchmark contains 2{,}500 documents, and each model is evaluated on the same document identifiers, giving 20{,}000 model-document extraction calls.

\begin{table}[t]
\centering
\scriptsize
\setlength{\tabcolsep}{4pt}
\renewcommand{\arraystretch}{1.05}
\begin{tabularx}{\columnwidth}{@{}l>{\raggedright\arraybackslash}Xl@{}}
\toprule
\textbf{Dataset} & \textbf{Entity types} & \textbf{Role} \\
\midrule
BC5CDR & chemical, disease & Main \\
NCBI Disease & disease & Main \\
BC2GM & gene & Main \\
JNLPBA & protein, DNA, RNA, cell line, cell type & Main \\
CHEMDNER & chemical & Robustness only \\
\bottomrule
\end{tabularx}
\caption{Five biomedical NER datasets. Each contributes 500 documents to the benchmark.}
\label{tab:datasets}
\end{table}

\subsection{Model panel}
The panel has eight models. Two are closed frontier systems (OpenAI GPT-5.5 and Claude Opus 4.7), and six are local open weight models covering dense, MoE, instruction-tuned, and reasoning families \citep{meta2024llama33,yang2025qwen3,deepseekai2025r1,minimax2025m2,mistral2025small32,abdin2024phi4}. The panel is used as a heterogeneous candidate source rather than a leaderboard of individual LLMs. Diversity across providers and model families creates the agreement and disagreement structure used by the downstream scorer. Local serving uses vLLM with an OpenAI-compatible endpoint \citep{kwon2023vllm}. The full per model panel table is in Appendix Table~\ref{tab:models}.

\subsection{Candidate master table}
\label{sec:candmaster}
For every document, we align model predictions into occurrence-level candidate rows under \texttt{(normalized\_text, type)} keys. Outputs assigned to the same occurrence slot are merged into one row with a vote bitmap, provider list, and surface variants.

Repeated mentions are retained as separate rows. If a document contains four mentions of the same chemical and the panel emits two surface variants, the candidate master keeps occurrence level rows rather than collapsing all evidence into one normalized string. This design matches the curator-facing unit of review.

Surface variants and provider offsets are retained for downstream analysis. Exact span labels use the first available provider offset. The \texttt{offset\_available\_count} and \texttt{offset\_ambiguity\_count} fields flag missing or many to many offset evidence for deterministic span alignment.

Multiset coverage caps repeated candidate keys at the corresponding gold occurrence count. Extra repeated candidates are overflow and cannot increase corpus recall. Gold annotations define training and evaluation labels. They are not used as input features at inference time. The resulting table has 23{,}700 rows and is the unit of analysis for all candidate scoring experiments.

\paragraph{Candidate universe.} The panel defines the recoverable candidate universe. All candidate-level scoring, coverage, and selection metrics in this paper are computed over that universe. Gold mentions missed by all panel models remain outside it. A post hoc scorer can prioritize surfaced candidates, but it cannot select candidates no source produced.

\paragraph{Two recall spaces.} BioConCal ranks only surfaced candidates. We therefore report $R_{\mathrm{cand}}$ over positive candidate rows. We also report $R_{\mathrm{corpus}}$ over all corpus gold mentions. This second metric includes the candidate generator ceiling because gold mentions that no panel model surfaced cannot be selected by a scorer.

Two ceiling conventions distinguish triage from source comparison. The row label panel ceiling is the maximum $R_{\mathrm{corpus}}$ under the paper's candidate row labels, with 2{,}124 positive candidate rows divided by 2{,}404 gold mentions, or 0.883. The stricter multiset source ceiling measures represented gold occurrences after capping repeated \texttt{(document, type, normalized\_text)} keys by gold occurrence counts, giving 2{,}057/2{,}404 = 0.856 for LLM-8. We use the stricter ceiling when comparing candidate generators such as BERN2.
Sampled documents absent from the candidate master have zero gold mentions in the released samples, so the corpus denominator includes all gold mentions in the sampled test documents and is unchanged by excluding zero-candidate, zero-gold documents from candidate-row scoring.

\subsection{Matching variants}
\label{sec:matching}
We use four matching variants to separate candidate identity from localization sensitivity. \textbf{Exact span type} declares a prediction correct when document, entity type, surface text, start, and end agree. \textbf{Exact surface type} drops the offset constraint. \textbf{Normalized text type} additionally tolerates case, whitespace, and punctuation variants and is the primary candidate label. \textbf{Relaxed overlap type} additionally accepts substring containment of normalized text or overlapping offsets of the same type. BioConCal predicts the normalized text type label. Exact surface and exact span labels are used to characterize downstream localization sensitivity in Section~\ref{sec:posthoc}.

\paragraph{Candidate keying.} The keying analyses test whether the benchmark conclusions depend on row granularity or label strictness. The main benchmark uses occurrence-aware rows and normalized text-type labels. Appendix Table~\ref{tab:keying_granularity_app} varies row aggregation, while Appendix Table~\ref{tab:keying_app} varies the correctness label with the candidate set held fixed.

\section{BioConCal for Candidate Triage}
\label{sec:bioconcal}
BioConCal is a supervised scorer for fixed panel surfaced candidate rows. For each row, it estimates correctness under the normalized text type label defined in Section~\ref{sec:matching}. Scores rank the candidate queue for curator review within the candidate universe defined by the panel. BioConCal therefore changes review priority, not candidate generation or exact offset localization. 

The input representation uses only features available at inference time. The features fall into four groups. \textbf{Agreement structure}: total, closed-model, and open-model agreement counts, plus a model-family diversity score over six families. \textbf{Mention features}: surface length, token count, document occurrence count, and binary indicators for dash, slash, parenthesis, digit, Greek letter, and capitalized abbreviation. \textbf{Surface availability}: exact and normalized surface presence in the document, plus the count and ambiguity of provider-supplied offsets. \textbf{Document features}: document length and one-hot entity type. Gold annotations define the supervised training label. Gold-derived signals such as gold entity count, hardness score, and difficulty bucket are excluded from the inputs to prevent leakage. The complete feature list is in Appendix Table~\ref{tab:feat_def}.

We instantiate BioConCal with two standard supervised scorers. Logistic regression gives a linear treatment of standardized features. A gradient boosted tree (GBT) scorer captures nonlinear interactions using \texttt{HistGradientBoostingClassifier} with 300 iterations, learning rate 0.06, and L2 regularization 0.5. We also include isotonic regression on the raw agreement count as a single feature calibration baseline. Expected calibration error (ECE) is computed with 10 equal-width bins. Calibration refers to empirical probability scoring and validation-selected operating points. It is not a distribution-free precision guarantee.

\subsection{Training and evaluation}
\label{sec:bioconcal_eval}
All train, validation, and test splits are at document level. The primary in-domain protocol uses a 60/20/20 split stratified by dataset, seed 42. We assert split integrity by checking that no $(\text{dataset},\text{doc\_id})$ pair appears in more than one split. The check is saved in \texttt{tables/split\_integrity\_check.json}.

We use a leave one dataset out (LODO) protocol to separate ranking transfer from operating point transfer. In each fold, BioConCal trains on three of the four main datasets and tests on the held-out dataset. Within the training pool, documents are split 75/25 into inner training and inner validation folds. CHEMDNER remains a separate robustness setting.

Operating points are selected only on validation data. For the 0.90, 0.95, and 0.97 precision targets, thresholds are selected on the validation fold and then \emph{frozen} before scoring the test fold. We never select thresholds on test data. We report AUROC, AUPRC, Brier score, and ECE, plus empirical test precision, test recall, selected count, and false positive count at each validation selected threshold.

\section{Experimental Setup}
\label{sec:setup}
The extraction protocol defines the candidate stream before BioConCal scoring. All panel models receive the same task instruction for a given dataset. The prompt asks each model to extract only explicitly mentioned biomedical entities from the dataset allowed type set and to return a JSON object with an \texttt{entities} array. Raw outputs are parsed into a common schema. For MiniMax-M2, Mistral-Small 3.2, and Phi-4 we use vLLM guided decoding with a JSON schema, which prevents long prose reasoning and constrains output to schema-valid JSON. Missing or parse-failed rows count as zero predictions. This keeps the candidate universe tied to surfaced model outputs rather than post hoc repair.

\section{Main Results}
\subsection{Candidate sources and agreement baselines}
We report single-model scores only to characterize the panel as a heterogeneous candidate source. The objective is not to rank LLM extractors, but to study the correctness of the candidates they surface. Among single models, OpenAI gives the highest normalized mention F1 (0.800) with the highest recall, while Claude gives higher precision (0.827) at lower recall. The open-weight models occupy distinct points in the precision-recall plane, which is the diversity the agreement analysis needs. The obvious way to convert a panel into a review queue is to threshold agreement count, and this section tests that heuristic directly. Voting baselines follow the standard ensemble pattern. Union maximizes recall but admits many false positives, higher agreement thresholds raise precision and cut recall, and unanimous voting reaches the highest precision (0.940) while recovering only 11.5 percent of gold mentions. The full single-model and voting tables are in Appendix Tables~\ref{tab:single} and~\ref{tab:voting}.

\subsection{Agreement is informative but incomplete}
Per-mention precision rises monotonically with agreement count, from 0.266 at $k{=}1$ to 0.940 at $k{=}8$, and the largest single-step gain is between $k{=}1$ and $k{=}2$. Even unanimous agreement does not eliminate error. The unanimous set contains 1{,}496 predictions of which 90 are false positives, a residual rate of 6.0 percent. This is the empirical form of the deceptive-simplicity problem. A candidate can be salient enough for every model to emit, yet still violate corpus-specific correctness through boundary, type, multiplicity, or annotation-convention mismatch. We inspect these high-agreement errors in the curation-oriented audit analysis. The per-$k$ calibration curve is in Appendix Figure~\ref{fig:agreement}.

\subsection{Candidate scoring beyond agreement count}
Learned candidate scoring converts panel agreement into a higher-yield review queue within the fixed candidate universe. Under the in-domain document-level split, BioConCal-GBT raises ranking AUROC from 0.753 for the one-feature agreement-count scorer to 0.910. It also lowers Brier score from 0.205 to 0.119. Because BioConCal scores only surfaced candidates, these gains are triage gains within the candidate universe rather than new candidate-generation recall. The full scorer comparison is in Table~\ref{tab:strongcal}, and the operating-point comparison is in Table~\ref{tab:risk}.

\begin{figure*}[t]
\centering
\includegraphics[width=\linewidth]{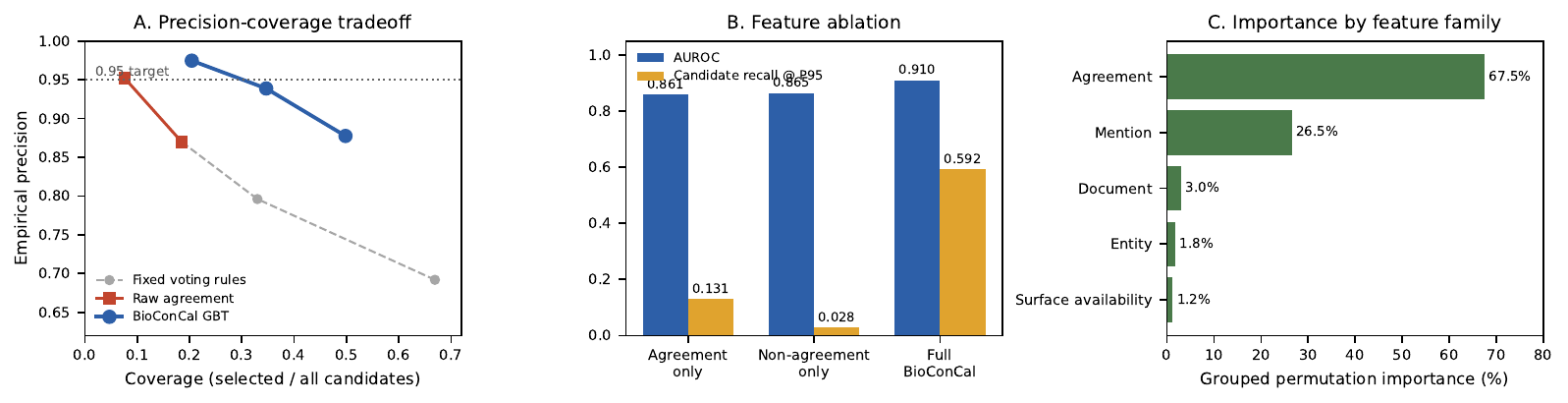}
\caption{Why learned candidate scoring improves over agreement count, on the document-level 60/20/20 test fold. (A) BioConCal improves the raw-agreement precision-coverage tradeoff and reaches far higher coverage at the validation-frozen 0.95 precision target. (B) Agreement carries the largest signal, while candidate metadata helps convert agreement into high-yield review selection. Agreement-only and non-agreement-only scoring each recover little precision-targeted candidate recall, while the main BioConCal model reaches 0.592. (C) Grouped permutation importance. Agreement features account for 67.5 percent of grouped importance, and mention features contribute 26.5 percent. Source: \texttt{scripts/main\_candidate\_scoring\_evidence.py}.}

\label{fig:evidence}
\end{figure*}

\subsection{Validation-targeted high-precision selection}
\begin{table*}[t]
\centering
\footnotesize
\begin{tabular}{lrrrrrr}
\toprule
Method & Sel. & FP & Test P & R$_{\mathrm{cand}}$ & R$_{\mathrm{corpus}}$ & Ceil. \\
\midrule
\multicolumn{7}{l}{\emph{Fixed voting rules (no threshold tuning)}}\\
Agreement $\geq$ 2 & 2{,}587 & 797 & 0.692 & 0.843 & 0.745 & 0.883 \\
Agreement $\geq$ 4 & 1{,}275 & 260 & 0.796 & 0.478 & 0.422 & 0.883 \\
Agreement $\geq$ 6 & 715 & 93 & 0.870 & 0.293 & 0.259 & 0.883 \\
Unanimous ($k=8$) & 293 & 14 & 0.952 & 0.131 & 0.116 & 0.883 \\
OpenAI $\cup$ Claude (closed $\geq 1$) & 2{,}743 & 743 & 0.729 & 0.942 & 0.832 & 0.883 \\
\midrule
\multicolumn{7}{l}{\emph{Validation-frozen P95 threshold}}\\
Raw agreement / 8 & 293 & 14 & 0.952 & 0.131 & 0.116 & 0.883 \\
Isotonic on agreement & 293 & 14 & 0.952 & 0.131 & 0.116 & 0.883 \\
BioConCal (logistic) & 1{,}229 & 85 & 0.931 & 0.539 & 0.476 & 0.883 \\
\textbf{BioConCal (GBT)} & \textbf{1{,}340} & \textbf{82} & \textbf{0.939} & \textbf{0.592} & \textbf{0.523} & \textbf{0.883} \\
\bottomrule
\end{tabular}
\caption{Validation-targeted high-precision selection on the document-level 60/20/20 test fold. Threshold-tuned methods use a 0.95 precision threshold selected on validation and applied frozen at test. $R_{\mathrm{cand}}$ is over 2{,}124 gold-positive candidate rows. $R_{\mathrm{corpus}}$ is over 2{,}404 corpus gold mentions in the row-label recall space. ``Ceil.'' is the within-panel row-label ceiling 2{,}124/2{,}404=0.883, distinct from the stricter multiset generator-source ceiling in Appendix Table~\ref{tab:gencompare}. Source: \texttt{tables/risk\_control\_selection\_docsplit.csv} and \texttt{tables/end\_to\_end\_interpretation.csv}.}
\label{tab:risk}
\end{table*}

The ranking gain translates into a larger high-precision review queue. At a 0.95 precision threshold selected on the validation fold and applied frozen at test, BioConCal-GBT selects 1{,}340 candidates with 82 false positives and empirical test precision 0.939. Raw agreement selects 293 candidates with 14 false positives at empirical precision 0.952. Within the candidate universe, BioConCal increases recall from 0.131 to 0.592 at comparable empirical precision.

Against all corpus gold mentions, the same operating point reaches 0.523 recall in the row-label recall space, below the panel-union ceiling of 0.883. Under the stricter source-comparison accounting used for generator and hybrid analyses, the same LLM-8 operating point is reported as 0.522 in Appendix Table~\ref{tab:failmodes}. The gain is therefore a curator-triage gain over surfaced candidates, not an end-to-end extraction recall gain. Compared with unanimity, BioConCal exposes 979 additional true candidates at the cost of 68 additional false positives. This is validation-targeted selection rather than a precision guarantee, so we report empirical test precision directly.

Monotonic transforms of the agreement count, including Platt scaling and isotonic regression, yield the same selected set as the raw agreement threshold. Split-conformal-inspired score cutoffs over the validation-fold false-positive quantile recover more candidates than raw thresholding but fall slightly below the 0.95 nominal precision on test. None of these baselines closes the precision-coverage gap achieved by BioConCal. The full table is in Appendix Table~\ref{tab:selective_app}.

\subsection{Stronger candidate-scoring baselines}
\label{sec:strongbaselines}
To test whether the gain comes from agreement count alone, vote identity alone, or candidate metadata alone, Table~\ref{tab:strongcal} compares progressively richer inference-time gold-free scorers under the same document-level split and validation-frozen P95 protocol.

\begin{table*}[t]
\centering
\scriptsize
\begin{tabular}{lrrrrrr}
\toprule
Scorer & AUROC & Brier & ECE & P@95 & R$_{\mathrm{cand}}$@95 & R$_{\mathrm{corpus}}$@95 \\
\midrule
Raw agreement count & 0.753 & 0.205 & 0.077 & 0.952 & 0.131 & 0.116 \\
Vote-vector logistic & 0.864 & 0.148 & 0.059 & 0.941 & 0.151 & 0.134 \\
Aggregate-agreement & 0.861 & 0.148 & 0.056 & 0.952 & 0.131 & 0.116 \\
Non-agreement only & 0.865 & 0.151 & 0.043 & 0.952 & 0.028 & 0.025 \\
BioConCal (logistic) & 0.891 & 0.135 & 0.037 & 0.931 & 0.539 & 0.476 \\
\textbf{BioConCal (GBT)} & \textbf{0.910} & \textbf{0.119} & \textbf{0.032} & \textbf{0.939} & \textbf{0.592} & \textbf{0.523} \\
\bottomrule
\end{tabular}
\caption{Stronger candidate-scoring baselines on the document-level 60/20/20 test fold. Vote-vector logistic uses the 8 per-model vote indicators. Aggregate-agreement uses the 4 agreement-structure features. Non-agreement only uses mention, surface, and document features. P@95, R$_{\mathrm{cand}}$@95, and R$_{\mathrm{corpus}}$@95 are at the validation-frozen P95 threshold. Source: \texttt{tables/stronger\_candidate\_scoring\_baselines\_p95.csv}. The complete logistic-and-GBT grid is in Appendix Table~\ref{tab:featabl_app}.}
\label{tab:strongcal}
\end{table*}

Three controls clarify the source of the gain. Scorers built only on agreement structure, whether the 8-way vote vector or the 4 aggregate-agreement features, reach AUROC near 0.86 but recover little precision-targeted recall. The agreement pattern alone is therefore informative but insufficient for high-yield review. On the primary split, a non-agreement scorer using mention, surface, and document features reaches AUROC 0.865 but selects only 63 rows at the validation-frozen P95 threshold. Candidate metadata alone therefore does not explain the main review-yield result.

BioConCal combines agreement and candidate metadata and reaches R$_{\mathrm{cand}}$ 0.592 at empirical precision 0.939. Permutation importance on the validation fold attributes about 67 percent of grouped importance to agreement features and 27 percent to mention features, with \texttt{closed\_agreement\_count} the strongest individual feature. The GBT and logistic forms of BioConCal are close in AUROC, 0.910 against 0.891. BioConCal is therefore best read as a panel-surfaced candidate verification framework rather than a single fixed classifier architecture.

A feature-removal ablation shows AUROC within $\pm 0.004$ of the main BioConCal model across variants that keep both agreement and non-agreement features. Only agreement-only and vote-pattern-only variants lose the high-yield selection behavior. Full ablations are in Appendix~\ref{app:featrob}.

\subsection{Robustness across splits}
\label{sec:robustness}

Across 10 document-level splits, BioConCal-GBT remains stable and consistently outperforms raw agreement, with AUROC 0.907$\pm$0.010. A document-level bootstrap on seed 42 gives BioConCal-GBT AUROC and R$_{\mathrm{cand}}$ 95\% intervals that do not overlap raw agreement. The full seed-wise and bootstrap results are in Appendix~\ref{app:robustness}.

\subsection{Cheaper-panel and single-model candidate baselines}
\label{sec:cheap_panel}
Panel size changes both scoring information and candidate coverage. We train candidate classifiers under restricted candidate universes on the same document-level 60/20/20 split and the same inference-time gold-free feature set. Each panel defines its own candidate universe by recomputing agreement features over the subset and dropping rows with zero in-panel votes. For comparability, R$_{full}$@95 is measured over the full eight-panel positive candidate set. Smaller panels are therefore penalized for candidates they cannot surface.

\begin{table}[t]
\centering
\scriptsize
\setlength{\tabcolsep}{2pt}
\begin{adjustbox}{max width=\columnwidth}
\begin{tabular}{lrrrr}
\toprule
Panel & $|M|$ & Cand. & AUROC & R$_{full}$@95 \\
\midrule
OpenAI & 1 & 12{,}628 & 0.815 & 0.170 \\
Claude & 1 & 9{,}324 & 0.814 & 0.276 \\
Best open & 1 & 7{,}979 & 0.880 & 0.262 \\
Closed-2 & 2 & 13{,}698 & 0.806 & 0.286 \\
Diverse-3 & 3 & 15{,}156 & 0.880 & 0.446 \\
Open-6 & 6 & 14{,}474 & 0.908 & 0.335 \\
Full-8 & 8 & 18{,}812 & 0.910 & 0.592 \\
\bottomrule
\end{tabular}
\end{adjustbox}
\caption{Cheap-panel baselines on the document-level test split. Cheaper panels recover much of AUROC, while the full panel gives higher high-precision recall by expanding the candidate universe. R$_{full}$@95 is empirical test recall over the full eight-panel positive candidate set at the validation-selected P95 operating point. Best open is MiniMax-M2. Diverse-3 is OpenAI plus MiniMax-M2 plus DeepSeek-R1. The full table with precision, FP, and coverage is in Appendix Table~\ref{tab:cheap_panel_app}.}
\label{tab:cheap_panel}
\end{table}

Cheaper panels recover much of the in-domain ranking AUROC, but they do not recover the same high-precision recall. Open-6 reaches AUROC 0.908 and Diverse-3 reaches 0.880, compared with 0.910 for Full-8. At the validation-frozen 0.95 precision target, single-model and closed-only pipelines reach at most 0.286 R$_{full}$@95. Diverse-3 reaches 0.446, while Full-8 reaches 0.592 by expanding the candidate universe. These results support cost-aware panel selection rather than a single universal panel. The appropriate panel depends on whether deployment prioritizes cost, reproducibility, or maximum recoverable review yield. The gain decomposition in Appendix Table~\ref{tab:gain_decomp_app} provides the corresponding feature-level comparison.

\subsection{LODO and OOD deployment conditions}
\label{sec:lodo_failure}
\begin{table}[t]
\centering
\scriptsize
\setlength{\tabcolsep}{2pt}
\begin{adjustbox}{max width=\columnwidth}
\begin{tabular}{lrrrr}
\toprule
Method & AUROC & Brier & P@95 & R$_{\mathrm{cand}}$@95 \\
\midrule
Raw agreement / 8 & 0.791 & 0.199 & 0.893 & 0.230 \\
Isotonic on agreement & 0.791 & 0.226 & 0.893 & 0.230 \\
BioConCal (logistic) & 0.856 & 0.151 & 0.898 & 0.170 \\
BioConCal (GBT) & 0.851 & 0.182 & 0.910 & 0.170 \\
\bottomrule
\end{tabular}
\end{adjustbox}
\caption{LODO macro mean across the four main datasets, with the threshold selected on the inner-validation fold and frozen at test. AUROC measures ranking transfer, while P@95 and R$_{\mathrm{cand}}$@95 depend on transferring a source-selected operating point. Values are the macro means from \texttt{tables/bioconcal\_lodo\_nested\_thresholds.csv}. Per-fold values are in Appendix Table~\ref{tab:lodo_fail_app}.}
\label{tab:lodo}
\end{table}
LODO separates ranking transfer from operating-point transfer. BioConCal improves mean AUROC over raw agreement, from 0.791 to 0.851 for GBT and 0.856 for logistic. At the validation-frozen P95 operating point, however, source-selected thresholds fall below the nominal precision target on average. These results argue for target-domain threshold validation before deployment. They do not imply that a small target calibration set always restores a high-yield operating point under entity-type shift. Per-fold transfer and target-domain recalibration results are in Appendix Tables~\ref{tab:lodo_fail_app} and~\ref{tab:lodo_recal_app}.

\paragraph{Deployment.} Deployment should select thresholds on target-domain validation data, report empirical precision after transfer, and use scores for triage rather than automatic insertion. The full deployment checklist is in Appendix~\ref{app:deploy}.

The reliability diagram in Appendix Figure~\ref{fig:reliability_app} shows that BioConCal predicted probabilities track empirical correctness closely on the in-domain fold.

\subsection{An external extractor as a candidate source}
\label{sec:bern2}
External candidate sources raise the recoverable ceiling, but they also change the verification burden. We evaluate BERN2 \citep{sung2022bern2} as an external candidate source on the same documents. Under strict multiset accounting, the ceiling is 0.856 for LLM-8, 0.737 for BERN2, and 0.901 for LLM-8\,$\cup$\,BERN2.

BioConCal-Hybrid addresses scorer comparability with a learned GBT over a merged LLM-8\,$\cup$\,BERN2 universe. It reaches AUROC 0.935 and AUPRC 0.948, and selects 1{,}847 candidate rows, including 1{,}716 row-label positives and 131 false positives. Under multiset corpus accounting, 1{,}705 selected gold occurrences over 2{,}404 corpus gold mentions give $R_{\mathrm{corpus}}=0.709$ at empirical precision 0.929. This recovers more corpus gold than the primary LLM-8 BioConCal workflow, but at lower empirical precision and with more false positives. The comparison supports the workflow decomposition. Generators set the recoverable universe, while learned verification determines review yield and false-positive burden within that universe. Full comparisons are in Appendix Tables~\ref{tab:gencompare} and~\ref{tab:hybrid_scorer_app}.

\section{Curation-Oriented Analysis}
\subsection{Span reconciliation after candidate triage}
\label{sec:posthoc}
Candidate triage resolves normalized entity identity, but curation still requires exact character localization. We therefore treat span reconciliation as a downstream step rather than as part of the scorer. Deterministic realignment searches exact, normalized, and supporting-span fallbacks, then chooses the match nearest the model offset. After alignment, exact-span F1 rises from 0.076 to 0.777 for OpenAI and from 0.162 to 0.708 for Claude. These gains show that candidate verification and localization are separable stages. The scorer prioritizes which normalized candidates should be reviewed, and alignment prepares selected candidates for offset-level inspection. Per-rule results are in Appendix Table~\ref{tab:span}.

\subsection{Auditing agreement errors and review boundaries}
\label{sec:audit}
Audit sheets make the residual review burden explicit. We release the 90 unanimous false positives and three disagreement audit sheets covering 758 further candidates (Appendix~\ref{app:audit}). These sheets surface cases where agreement, source identity, or normalization cues may diverge from corpus convention. Their category labels are auto-suggested and are intended for follow-up annotation review, not as manually verified error labels.

Stratified calibration preserves the same broad increasing trend, with expected deviations in small dataset and entity type strata. The normalization ablation preserves the central pattern across matching variants (Appendix Figure~\ref{fig:cal_by_ds_app} and Appendix~\ref{app:norm_app}). These checks define the boundary of the review workflow rather than a new automatic decision rule. BioConCal produces the ranked candidate queue, span reconciliation prepares selected normalized candidates for offset inspection, audit sheets expose cases requiring human adjudication, and source expansion changes the precision coverage tradeoff. Appendix Table~\ref{tab:failmodes} makes this boundary quantitative. The primary LLM-8 BioConCal workflow gives the largest high precision review yield in the LLM-8 queue, while BioConCal-Hybrid recovers more corpus gold at lower empirical precision.

\section{Conclusion}
We introduced a candidate-level panel output benchmark for panel surfaced candidate verification, a curator triage layer between biomedical entity extraction and human review. In the in-domain document split, BioConCal raised AUROC from 0.753 for raw agreement to 0.910. At the validation selected 0.95 precision target, it selected 1{,}340 candidate rows at empirical precision 0.939, compared with 293 rows for raw agreement. This operating point gives $R_{\mathrm{cand}}=0.592$ and row label $R_{\mathrm{corpus}}=0.523$ within a panel universe whose row label ceiling is 0.883. The results support a workflow decomposition in which candidate sources set the recoverable universe, learned scoring determines review yield within that universe, and source expansion through BERN2 trades coverage for lower empirical precision. Target domain validation, deterministic span reconciliation, and human review remain required before selected candidates can support curated resource updates.

\section*{Limitations}
\paragraph{Candidate universe and verifier scope.} BioConCal scores candidates that have already been surfaced by the panel. It does not generate new mentions and cannot recover gold mentions missed by every panel model. Within panel corpus recall is therefore bounded by the row label ceiling of 0.883 on the test fold. BioConCal is also not an upper bound on context-aware verification. Richer context-aware verifiers can be used after BioConCal when a candidate needs textual evidence beyond panel and candidate metadata.

\paragraph{Portability and operating points.} BioConCal is tied to the candidate sources, prompts, decoding constraints, corpus convention, and entity type mix used to train and validate it. If any of these change, the classifier and its operating threshold require target domain validation before deployment. The LODO and target recalibration analyses show why this validation matters under entity type shift, but they do not guarantee that 25 to 100 labeled documents will recover the same yield in every target corpus. Prompt sensitivity is evaluated on an open weight subset in Appendix~\ref{app:promptsens}.

\paragraph{Human review and reproducibility.} BioConCal scores define a review priority, not automatic acceptance or rejection. False positives can contaminate curated biomedical resources if accepted without review, and false negatives can leave useful candidates outside the high-priority queue. High scores, low scores, and audit categories should therefore be treated as triage signals for human adjudication. The released audit categories are auto-suggested and are intended to guide follow-up review rather than serve as verified error labels. Closed API providers introduce cost and reproducibility asymmetry, while the open weight six model variant is the reproducible alternative. CHEMDNER is kept as a robustness dataset because its public mirror lacks reliable character offsets. We report empirical test precision and do not claim distribution-free precision guarantees.

\section*{Ethics and Reproducibility}
All datasets are public biomedical NER resources, and the work introduces no new human-subject data. We release all analysis scripts, and the open-weight six-model variant is locally reproducible. The full eight-model benchmark depends on closed API access. Reproduction commands are in Appendix~\ref{app:repro}.

\bibliography{custom}

\appendix
\raggedbottom
\setlength{\textfloatsep}{6pt plus 2pt minus 2pt}
\setlength{\floatsep}{6pt plus 2pt minus 2pt}
\setlength{\intextsep}{6pt plus 2pt minus 2pt}
\setlength{\abovecaptionskip}{3pt}
\setlength{\belowcaptionskip}{2pt}
\section{Reproducibility commands}
\label{app:repro}
The candidate master and downstream analysis tables can be regenerated from the released sample documents, parsed provider outputs, consensus candidate artifacts, and table sources. Full extraction regeneration additionally requires rerunning the provider calls; the downstream scripts below are run from the project root.

\noindent
\begin{minipage}{\linewidth}\small
\texttt{build\_candidate\_master.py}: candidate-level master.\\
\texttt{train\_bioconcal\_v3.py}: doc-level 60/20/20 + LODO with inner-val thresholds.\\
\texttt{selective\_baselines.py}: selective and conformal baselines.\\
\texttt{panel\_composition\_ablation.py}: panel ablations.\\
\texttt{feature\_importance.py}: permutation importance + figure.\\
\texttt{span\_alignment\_audit\_sample.py}: 100-row span audit sample.\\
\texttt{extended\_span\_alignment.py}: span alignment metrics.\\
\texttt{audit\_unanimous\_fps.py}: unanimous FP audit sheet (multiset).\\
\texttt{finalize\_unanimous\_fp\_audit.py}: refined audit with subcategories.\\
\texttt{disagreement\_audit.py}: candidate audit sheets.\\
\texttt{agreement\_calibration\_by\_strata.py}: per-stratum.\\
\texttt{normalization\_ablation.py}: normalization deltas.\\
\texttt{stronger\_candidate\_scoring\_baselines.py}: consolidated stronger-baseline table.\\
\texttt{main\_candidate\_scoring\_evidence.py}: 3-panel main evidence figure.\\
\texttt{train\_hybrid\_candidate\_scorer.py}: learned LLM-8\,$\cup$\,BERN2 candidate scorer.
\end{minipage}

Script locations are mixed by design. The candidate-master, training, ablation, and audit-generation scripts live under \texttt{paper\_experiment\_materials/\allowbreak strong\_accept/scripts/}. The unanimous-audit finalization, stratum calibration, main evidence, stronger-baseline, and hybrid-scoring scripts live under top-level \texttt{scripts/}.

\section{Related work, model panel, and baseline tables}
\label{app:baselines}
\paragraph{Biomedical NER and LLM candidate generation.}
Public biomedical NER benchmarks cover chemical, disease, gene, protein, nucleic acid, cell line, and cell type mentions \citep{li2016bc5cdr,dogan2014ncbi,smith2008biocreative,collier2004jnlpba,krallinger2015chemdner}. Biomedical encoders and generative biomedical language models include BioBERT, PubMedBERT, and BioGPT \citep{lee2020biobert,gu2021pubmedbert,luo2022biogpt}. Supervised biomedical NER work studies multi-dataset auxiliary learning, discontinuous and overlapping entities, and dataset availability debt \citep{watanabe2022auxiliary,khandelwal2022biomedical,fries2022dataset}. Established systems combine neural recognition with entity normalization, and BERN2 \citep{sung2022bern2} is a widely used biomedical NER and normalization reference point. LLM-based biomedical information extraction has been studied for zero-shot biomedical reasoning \citep{nagar2024llms}, LLM versus encoder biomedical NER tradeoffs \citep{obeidat2025llms}, flat and nested NER under generative frameworks \citep{lv2025unified}, and multi-LLM workflows for clinical knowledge graph construction \citep{das2026clinical}. These works ask whether an extractor or generator can produce entity predictions under a task schema. We start after candidate sources have produced candidates and evaluate aligned candidate rows as review decisions.

\paragraph{Panel-surfaced verification for curator review.}
Multi LLM consensus paired with selective human review has been studied as an annotation pattern \citep{yuan2025multillm}. BioConCal studies a different workflow layer. Given a fixed stream of candidates surfaced by a multi-model panel, the question is whether those candidates can be ranked for curator review using agreement structure and inference time gold-free candidate features. External extractors such as BERN2 can expand the candidate universe, but they do not remove the need for source-specific verification. Context-aware verification uses richer document evidence to adjudicate individual candidates. It is complementary to the panel-derived triage layer studied here, which first ranks a large candidate stream for review. Appendix Table~\ref{tab:external} contrasts evaluation units across related settings.

\paragraph{Agreement, calibration, and LLM uncertainty.}
Ensembles and self-consistency are widely used to improve reliability \citep{dietterich2000ensemble,wang2023selfconsistency}. Probabilistic calibration of classifier outputs is standard \citep{platt1999probabilistic,niculescu2005predicting}. Selective classification allows abstention on low confidence inputs \citep{geifman2017selective}, conformal prediction provides distribution-free coverage guarantees \citep{angelopoulos2021gentle}, and confident learning and dataset cartography use disagreement to identify possibly mislabeled examples \citep{northcutt2021confident,swayamdipta2020dataset}. For single-LLM uncertainty, \citet{lin2023uncertainty} quantifies black-box uncertainty via the semantic dispersion of sampled generations, and \citet{tian2023calibration} elicits verbalized confidence from RLHF-fine-tuned LLMs. In biomedical NLP, \citet{deoliveira2025calibration} reports that out-of-the-box LLM calibration is poor while post hoc calibration helps, and \citet{yuan2024chatgpt} documents systematic overconfidence in the black-box calibration of a closed LLM. Our work treats multi-LLM agreement as panel evidence of salience and learns a downstream scorer from agreement and candidate metadata. It targets validation-selected operating points for review triage rather than distribution-free guarantees or automatic curation.

\begin{table*}[t]
\centering
\footnotesize
\begin{tabular}{p{0.19\linewidth}p{0.12\linewidth}p{0.20\linewidth}ccp{0.15\linewidth}}
\toprule
Line of work & Unit of prediction & Main objective & Panel agr. & Triage score & Relation to BioConCal \\
\midrule
BERN2 \citep{sung2022bern2} & Mention span & End-to-end NER and normalization & No & No & Candidate source \\
Biomedical BERT and PubMedBERT NER \citep{lee2020biobert,gu2021pubmedbert} & Token and span & Supervised biomedical NER & No & No & Candidate source \\
Extractor-level NER evaluation & Model or span & Rank or score standalone extractors & No & No & Different layer \\
Multi-LLM consensus annotation \citep{yuan2025multillm} & Item label & Consensus plus human review & Yes (count) & No & Related, count only \\
BioConCal (this work) & Panel-surfaced candidate & Validation-targeted candidate triage & Yes & Yes & This work \\
\bottomrule
\end{tabular}
\caption{Evaluation units across related settings. BioConCal conditions on surfaced candidates and evaluates candidate-level triage rather than end-to-end extraction. ``Panel agr.'' is whether the method uses multi-LLM panel agreement structure, and ``Triage score'' is whether it emits a validation-targeted candidate-correctness score.}
\label{tab:external}
\end{table*}
\begin{table}[!htbp]
\centering
\footnotesize
\begin{tabular}{p{0.30\linewidth}p{0.18\linewidth}p{0.40\linewidth}}
\toprule
Model & Access & Type and size \\
\midrule
GPT-5.5 & Closed API & frontier, undisclosed \\
Claude Opus 4.7 & Closed API & frontier, undisclosed \\
Llama 3.3 70B & Local & open-weight dense, 70B \\
Qwen3-30B-A3B-Instruct & Local & open-weight MoE, 30.5B/3.3B active \\
DeepSeek-R1 32B & Local & open-weight reasoning, 32B \\
MiniMax-M2 & Local & open-weight MoE, 230B/10B active \\
Mistral-Small 3.2 24B & Local & open-weight dense, 24B \\
Phi-4 & Local & open-weight dense, 14B \\
\bottomrule
\end{tabular}
\caption{Eight-model panel. Six models are served locally with vLLM.}
\label{tab:models}
\end{table}

\begin{table}[!htbp]
\centering
\footnotesize
\begin{tabular}{lccc}
\toprule
Model & P & R & F1 \\
\midrule
OpenAI GPT-5.5 & 0.788 & 0.812 & 0.800 \\
Claude Opus 4.7 & 0.827 & 0.629 & 0.715 \\
MiniMax-M2 & 0.665 & 0.433 & 0.525 \\
Mistral-Small 3.2 & 0.659 & 0.386 & 0.487 \\
Qwen3-30B-A3B & 0.612 & 0.361 & 0.454 \\
DeepSeek-R1 32B & 0.743 & 0.319 & 0.446 \\
Phi-4 & 0.666 & 0.307 & 0.420 \\
Llama 3.3 70B & 0.731 & 0.283 & 0.408 \\
\bottomrule
\end{tabular}
\caption{Single-model normalized mention F1, overall micro on the four main datasets (excludes CHEMDNER).}
\label{tab:single}
\end{table}

\begin{table}[!htbp]
\centering
\footnotesize
\begin{tabular}{lccc}
\toprule
Rule & P & R & F1 \\
\midrule
Union, any model & 0.573 & 0.880 & 0.694 \\
OpenAI $\cup$ Claude & 0.763 & 0.854 & 0.806 \\
Agreement $\geq 2$ & 0.711 & 0.753 & 0.731 \\
Agreement $\geq 4$ & 0.805 & 0.428 & 0.559 \\
Agreement $\geq 6$ & 0.878 & 0.263 & 0.405 \\
Unanimous ($k=8$) & 0.940 & 0.115 & 0.205 \\
\bottomrule
\end{tabular}
\caption{Voting baselines on the 8-model panel. The closed-only union is the best F1 voting rule, but unanimous voting trades almost all recall for precision.}
\label{tab:voting}
\end{table}

\section{BERN2 NER reference point}
\label{app:bern2}
We run the BERN2 NER model (the recognition component of BERN2, without entity normalization) on the same 500-document-per-dataset samples, using the official \texttt{multi\_ner} module with the released \texttt{dmis-lab/bern2-ner} weights. BERN2 heads map to our schema as disease, drug to chemical, gene, DNA, RNA, cell line, and cell type, while species and mutation are ignored. BERN2 merges gene and protein into one head, so BC5CDR, NCBI Disease, and BC2GM use clean mappings while JNLPBA protein is approximate and reported separately. Table~\ref{tab:bern2res} reports normalized mention P/R/F1 and Table~\ref{tab:bern2overlap} the candidate-source overlap. Run details are in \texttt{bern2\_run\_report.md}.

\begin{table}[!htbp]
\centering
\footnotesize
\begin{tabular}{lrrr}
\toprule
Dataset & P & R & F1 \\
\midrule
BC5CDR & 0.874 & 0.772 & 0.820 \\
NCBI Disease & 0.869 & 0.875 & 0.872 \\
BC2GM & 0.864 & 0.868 & 0.866 \\
Overall (clean) & 0.873 & 0.782 & 0.825 \\
\midrule
JNLPBA (approx.) & 0.509 & 0.740 & 0.603 \\
\bottomrule
\end{tabular}
\caption{BERN2 NER model, normalized mention matching on the sampled documents. Overall is over BC5CDR, NCBI Disease, and BC2GM with clean type mappings. JNLPBA is approximate because BERN2 merges gene and protein. Source: \texttt{tables/bern2\_reference\_results.csv}.}
\label{tab:bern2res}
\end{table}

\begin{table*}[!t]
\centering
\footnotesize
\begin{tabular}{lrrrrr}
\toprule
Dataset & Gold & Panel covered & BERN2 covered & Panel ceiling & Panel+BERN2 ceiling \\
\midrule
BC5CDR & 9{,}965 & 9{,}117 & 8{,}307 & 0.915 & 0.939 \\
NCBI Disease & 506 & 460 & 446 & 0.909 & 0.953 \\
BC2GM & 636 & 499 & 556 & 0.785 & 0.936 \\
JNLPBA & 1{,}140 & 1{,}046 & 848 & 0.918 & 0.930 \\
\bottomrule
\end{tabular}
\caption{Candidate-source overlap. ``Panel covered'' and ``BERN2 covered'' are gold mentions whose normalized key is surfaced by each source. Adding BERN2 to the panel raises the candidate-universe ceiling, most on BC2GM genes. Source: \texttt{tables/bern2\_candidate\_source\_overlap.csv}.}
\label{tab:bern2overlap}
\end{table*}

Table~\ref{tab:gencompare} decomposes the pipeline into candidate-generation ceiling and triage yield on the same document-level test fold. The eight-LLM panel reaches strict multiset ceiling 0.856, BERN2 alone reaches 0.737, and the additive generator union reaches 0.901 by surfacing gold mentions the panel misses. The union row in Table~\ref{tab:gencompare} is an exploratory merge rule, not a learned scorer over a shared representation. Table~\ref{tab:hybrid_scorer_app} therefore adds BioConCal-Hybrid, a learned GBT over the merged LLM-8\,$\cup$\,BERN2 universe with LLM agreement features where available, BERN2 source indicators, and BERN2 confidence scores.

\begin{table*}[!t]
\centering
\footnotesize
\begin{tabular}{lrrcclccrr}
\toprule
Candidate source & Cands & Repr. & Ceiling & Src & Scorer & AUROC & P@0.95 & R$_{\mathrm{cand}}$ & R$_{\mathrm{corpus}}$ \\
 & & gold & & F1 & & & & & \\
\midrule
LLM-8 only & 3866 & 2057 & 0.856 & -- & BioConCal & 0.910 & 0.939 & 0.592 & 0.522 \\
BERN2 only & 2199 & 1772 & 0.737 & 0.770 & take-all & n/a & 0.806 & n/a & 0.737 \\
LLM-8 $\cup$ BERN2 & 6065 & 2166 & 0.901 & -- & merge & n/a & 0.836 & n/a & 0.550 \\
Diverse-3 only & 2968 & 1207 & 0.502 & -- & BioConCal & 0.880 & 0.944 & 0.490 & 0.269 \\
Diverse-3 $\cup$ BERN2 & 5167 & 1974 & 0.821 & -- & merge & n/a & 0.680 & n/a & 0.567 \\
\bottomrule
\end{tabular}
\caption{Candidate-generator comparison on the document-level test fold (seed 42). The generator fixes the candidate-universe \emph{ceiling} (Repr.\ gold $=$ multiset gold occurrences covered; ceiling $=$ Repr.$/2404$); BioConCal determines \emph{triage yield} (AUROC, P@0.95, R$_{\mathrm{cand}}$, R$_{\mathrm{corpus}}$) within that universe. Scorers are not forced to match: a single extractor has no agreement features, so BERN2-only is take-all (no triage) and union rows use an exploratory merge rule. ``Src F1'' is take-all extractor F1 on the test fold (single extractors only; ``--'' for panels, which have no single operating point). R$_{\mathrm{corpus}}$ is comparable across all rows; R$_{\mathrm{cand}}$ is within-universe and reported only for BioConCal-scored panels. JNLPBA gene/protein mapping is approximate. Source: \texttt{scripts/main\_candidate\_generator\_comparison.py}, \texttt{tables/main\_candidate\_generator\_comparison.csv}.}
\label{tab:gencompare}
\end{table*}

\begin{table*}[!t]
\centering
\scriptsize
\setlength{\tabcolsep}{2pt}
\begin{adjustbox}{max width=\textwidth}
\begin{tabular}{p{0.19\linewidth}p{0.14\linewidth}rrrrrrrrrrc}
\toprule
Workflow & Cand.\ source & Cands & Repr. & Ceiling & AUROC & AUPRC & Test P & Sel. & Row TP & FP & R$_{\mathrm{corpus}}$ & Fit \\
\midrule
LLM-8 BioConCal & LLM-8 & 3{,}866 & 2{,}057 & 0.856 & 0.910 & n/a & 0.939 & 1{,}340 & 1{,}258 & 82 & 0.522 & Primary \\
BERN2 take-all & BERN2 & 2{,}199 & 1{,}772 & 0.737 & n/a & n/a & 0.806 & 2{,}199 & 1{,}772 & 427 & 0.737 & Unfiltered \\
LLM-8 $\cup$ BERN2 merge & additive union & 6{,}065 & 2{,}166 & 0.901 & n/a & n/a & 0.836 & 1{,}583 & 1{,}323 & 260 & 0.550 & Unfiltered \\
BioConCal-Hybrid & merged union & 4{,}140 & 2{,}150 & 0.894 & 0.935 & 0.948 & 0.929 & 1{,}847 & 1{,}716 & 131 & 0.709 & Tradeoff \\
\bottomrule
\end{tabular}
\end{adjustbox}
\caption{Learned hybrid candidate scoring on the document-level test fold. BioConCal-Hybrid trains a GBT over a merged LLM-8\,$\cup$\,BERN2 candidate universe; BERN2-only rows have zero LLM votes and source indicators mark them as BERN2-only, while overlap rows retain LLM features and receive BERN2 indicators. Row TP counts selected candidate rows labeled positive. R$_{\mathrm{corpus}}$ uses multiset corpus accounting; for the hybrid row, 1{,}705 selected multiset gold mentions over 2{,}404 corpus gold mentions gives R$_{\mathrm{corpus}}=0.709$. The hybrid reduces the false-positive burden relative to the exploratory merge and recovers more true candidate rows than LLM-only BioConCal, but its empirical precision 0.929 is lower than the primary LLM-8 BioConCal precision of 0.939. This is a candidate-triage comparison, not a claim that BioConCal is a better extractor than BERN2. Source: \texttt{scripts/train\_hybrid\_candidate\_scorer.py}, \texttt{tables/hybrid\_candidate\_scorer\_results.csv}.}
\label{tab:hybrid_scorer_app}
\end{table*}

\section{Feature definitions}
\label{app:feat_def}
\begin{table*}[!t]
\centering
\scriptsize
\begin{tabular}{p{0.22\linewidth}p{0.14\linewidth}p{0.58\linewidth}}
\toprule
Feature & Group & Definition \\
\midrule
\texttt{agreement\_count} & Agreement & Number of providers (of 8) emitting the same $(\text{normalized text},\text{type})$ key. \\
\texttt{closed\_agreement\_count} & Agreement & Subcount over closed-API providers (OpenAI, Claude). \\
\texttt{open\_agreement\_count} & Agreement & Subcount over six open-weight providers. \\
\texttt{model\_family\_\allowbreak diversity\_score} & Agreement & Distinct families among agreeing providers (six families). \\
\texttt{mention\_length\_chars} & Mention & Character length of predicted surface. \\
\texttt{mention\_token\_count} & Mention & Whitespace-token count of predicted surface. \\
\texttt{mention\_occurrences\_\allowbreak in\_doc} & Mention & Literal occurrences of predicted surface in the source document. \\
\texttt{has\_*} indicators & Mention & Dash, slash, paren, digit, Greek letter, or capitalized abbreviation in the surface. \\
\texttt{exact\_surface\_in\_doc} & Surface availability & 1 if predicted surface is a literal substring of the document. \\
\texttt{normalized\_in\_doc} & Surface availability & 1 if normalized surface is a substring of normalized document. \\
\texttt{offset\_available\_count} & Surface availability & Providers that supplied a numeric offset. \\
\texttt{offset\_ambiguity\_count} & Surface availability & Distinct $(\text{start},\text{end})$ tuples across providers. \\
\texttt{doc\_text\_length} & Document & Character length of source document. \\
\texttt{entity\_type} & Entity & Unified eight-type schema, one-hot encoded. \\
\bottomrule
\end{tabular}
\caption{BioConCal gold-free input-feature definitions. Training labels come from gold annotations and are not listed here. Per-feature permutation importance is in Figure~\ref{fig:feat_imp}.}
\label{tab:feat_def}
\end{table*}

\section{Auto-suggested audit sheets}
\label{app:audit}
We release four candidate audit sheets. Set A is the unanimous false-positive list (90 candidates under multiset normalized matching). Set B is high-agreement false positives with $k \in \{6,7\}$ (358 candidates). Set C is minority true positives with $k \leq 2$ (200 sampled). Set D is closed-versus-open disagreement (200 sampled). All sheets are CSV with auto-suggested category labels and are released to support follow-up annotation review; the categories have not been manually verified. The final Set A source is the revision file \texttt{unanimous\_fp\_audit\_final.csv}; older combined audit artifacts are not used for the final 90-candidate count. A 100-row span-alignment audit sample stratified across repeated-surface, multi-candidate-span, long-or-nested, overlap, and random cases is in \texttt{audit/span\_alignment\_\allowbreak manual\_audit\_sample.csv}.

\section{Span-alignment method success rates}
\label{app:span_modes}
\begin{table*}[!t]
\centering
\footnotesize
\begin{tabular}{lrrrr}
\toprule
Alignment method & Total & With gold & Post=Gold & Rate \\
\midrule
Exact surface match & 12{,}628 & 12{,}402 & 9{,}637 & 0.777 \\
\bottomrule
\end{tabular}
\caption{Span-alignment method success rates on OpenAI predictions across the four main datasets. Source: \texttt{tables/span\_alignment\_error\_modes\_v2.csv}.}
\label{tab:span_modes}
\end{table*}

\section{Full feature ablation}
\label{app:featabl}
\begin{table*}[!t]
\centering
\footnotesize
\begin{tabular}{llrrrrrrr}
\toprule
Features & Clf. & \#feat. & AUROC & Brier & ECE & P@95 & R$_{\mathrm{cand}}$@95 & Sel \\
\midrule
A. agreement\_count only & logistic & 1 & 0.753 & 0.205 & 0.077 & 0.952 & 0.131 & 293 \\
A. agreement\_count only & GBT & 1 & 0.751 & 0.199 & 0.050 & 0.952 & 0.131 & 293 \\
B. vote-pattern only (8 indicators) & logistic & 8 & 0.864 & 0.148 & 0.059 & 0.941 & 0.151 & 341 \\
B. vote-pattern only (8 indicators) & GBT & 8 & 0.859 & 0.151 & 0.053 & 0.953 & 0.143 & 318 \\
C. aggregate agreement only (4 feats) & logistic & 4 & 0.861 & 0.148 & 0.056 & 0.952 & 0.131 & 293 \\
C. aggregate agreement only (4 feats) & GBT & 4 & 0.860 & 0.149 & 0.054 & 0.000 & 0.000 & 0 \\
D. non-agreement only (15 feats) & logistic & 15 & 0.829 & 0.169 & 0.054 & 0.913 & 0.158 & 367 \\
D. non-agreement only (15 feats) & GBT & 15 & 0.865 & 0.151 & 0.043 & 0.952 & 0.028 & 63 \\
E. agg.\ agree.\ + non-agree. & logistic & 19 & 0.891 & 0.135 & 0.037 & 0.931 & 0.539 & 1{,}229 \\
E. agg.\ agree.\ + non-agree. & GBT & 19 & 0.910 & 0.119 & 0.032 & 0.939 & 0.592 & 1{,}340 \\
F. vote-pattern + non-agreement & logistic & 23 & 0.892 & 0.134 & 0.036 & 0.930 & 0.545 & 1{,}245 \\
F. vote-pattern + non-agreement & GBT & 23 & 0.908 & 0.120 & 0.037 & 0.939 & 0.554 & 1{,}253 \\
G. all-feature diagnostic & logistic & 27 & 0.892 & 0.134 & 0.037 & 0.928 & 0.540 & 1{,}237 \\
G. all-feature diagnostic & GBT & 27 & 0.910 & 0.119 & 0.026 & 0.946 & 0.557 & 1{,}251 \\
\bottomrule
\end{tabular}
\caption{Complete feature-ablation grid (logistic + GBT, all seven feature sets). Row E is the main 19-feature BioConCal definition used in the paper; row G is an all-feature diagnostic that additionally includes the 8 per-model vote indicators. Row C-GBT (aggregate-agreement-only, GBT) selects zero candidates at the val-frozen P95 threshold because the GBT predicted-probability distribution on aggregate-agreement-only features has no validation point above the 0.95 precision target; this is a quirk of finite-sample threshold selection on a low-dimensional GBT and is included for completeness. Source: \texttt{tables/feature\_ablation\_baselines.csv}.}
\label{tab:featabl_app}
\end{table*}

\begin{figure*}[!t]
\centering
\includegraphics[width=0.85\linewidth]{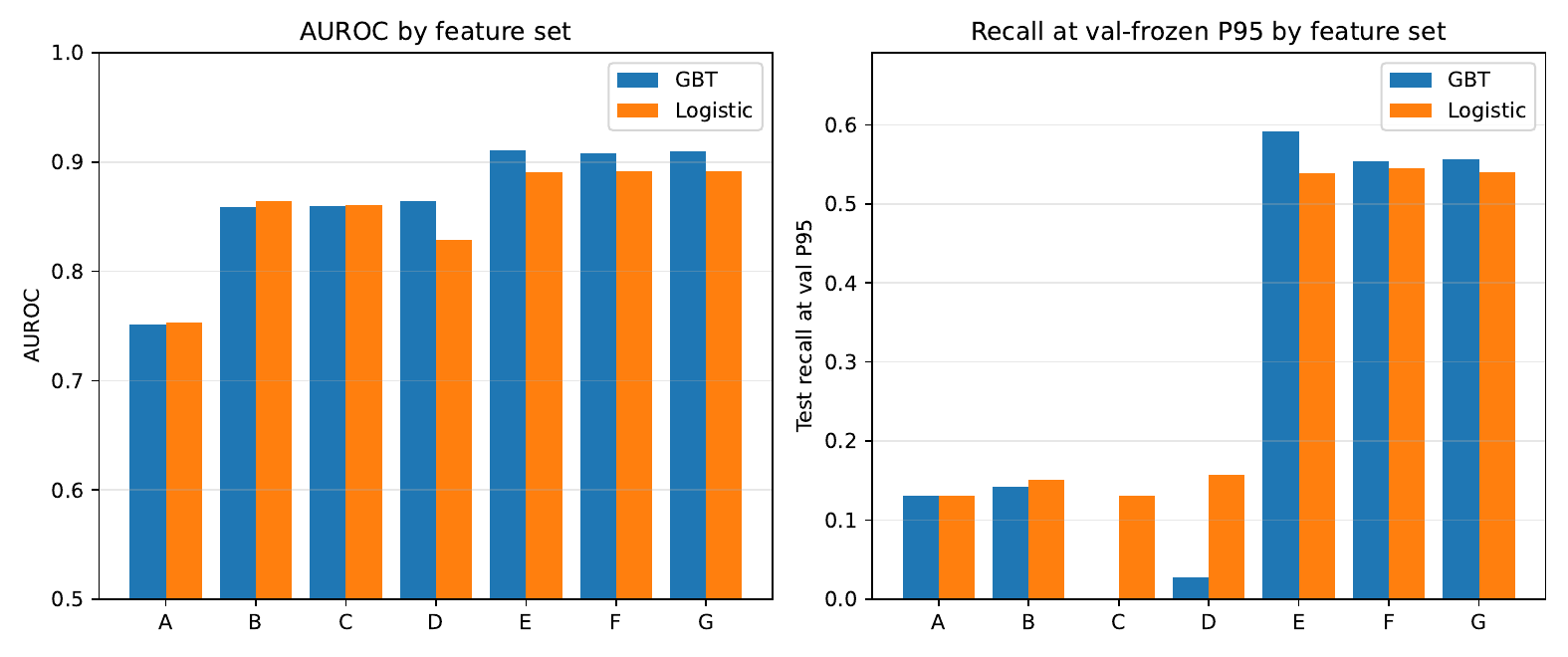}
\caption{Feature-ablation AUROC (left) and recall at the validation-frozen P95 threshold (right) for GBT and logistic classifiers.}
\label{fig:featabl}
\end{figure*}

\begin{figure*}[!t]
\centering
\includegraphics[width=0.85\linewidth]{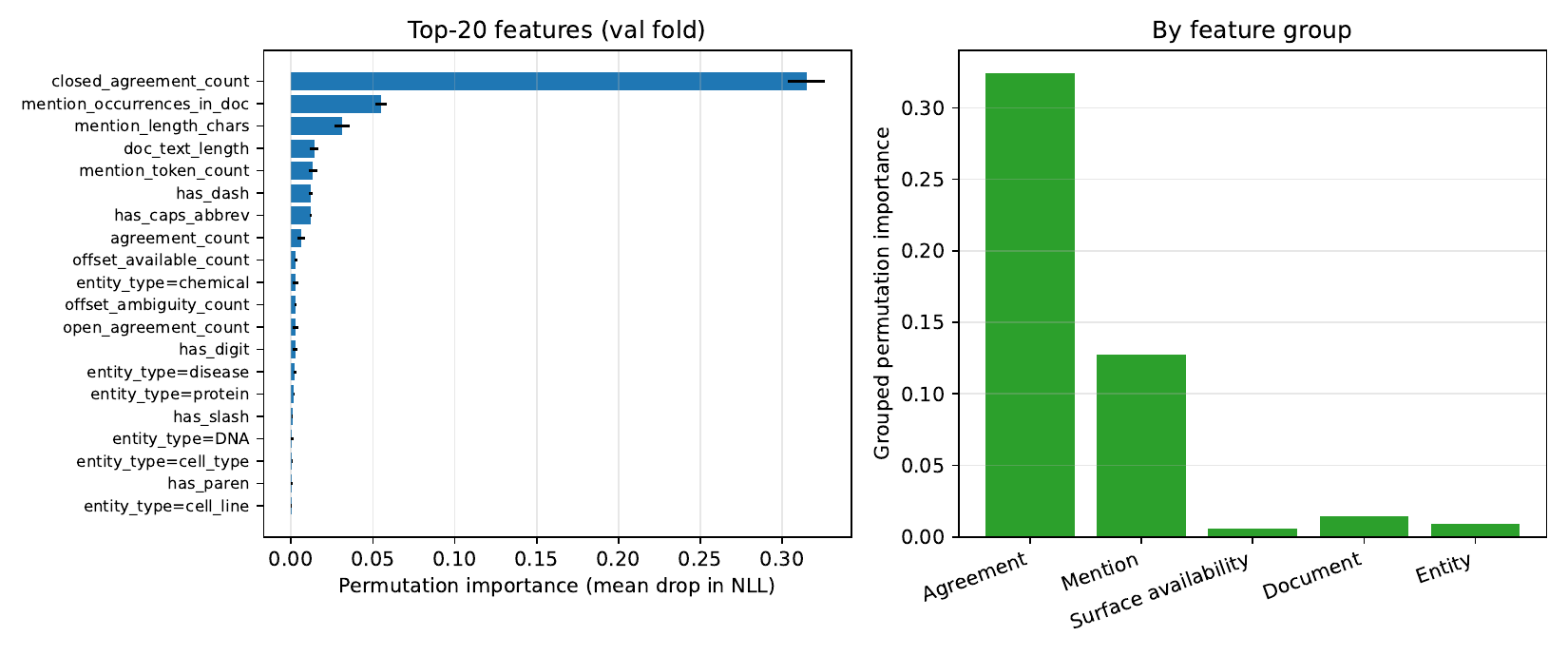}
\caption{Permutation feature importance for BioConCal-GBT, doc-level validation fold (mean drop in negative log-loss, five shuffles). Left: top-20 individual features. Right: grouped importance by feature family.}
\label{fig:feat_imp}
\end{figure*}

\section{Robustness: splits, bootstrap, and candidate keying}
\label{app:robustness}
\subsection{Per-seed and bootstrap detail}
Table~\ref{tab:robust_seedwise} gives the per-seed results for the headline scorer behind the summary in Table~\ref{tab:robust}. Table~\ref{tab:bootstrap_ci} gives the document-level bootstrap 95\% confidence intervals on the seed-42 test fold (2{,}000 resamples of documents, stratified by dataset; documents rather than candidates are resampled because within-document candidates are correlated). Both are produced without rerunning extraction, from \texttt{candidate\_master.csv}.

\begin{table*}[!t]
\centering
\footnotesize
\begin{tabular}{lccccc}
\toprule
Scorer & AUROC & P@0.95 & R$_{\mathrm{cand}}$@0.95 & Sel. & FP \\
\midrule
Raw agreement count & 0.742$\pm$0.010 & 0.931$\pm$0.034 & 0.149$\pm$0.044 & 356$\pm$114 & 28$\pm$23 \\
Isotonic on agreement & 0.741$\pm$0.010 & 0.931$\pm$0.034 & 0.149$\pm$0.044 & 356$\pm$114 & 28$\pm$23 \\
Vote-vector logistic & 0.852$\pm$0.011 & 0.898$\pm$0.077 & 0.043$\pm$0.076 & 102$\pm$176 & 9$\pm$16 \\
Aggregate-agreement scorer & 0.846$\pm$0.010 & 0.935$\pm$0.029 & 0.144$\pm$0.035 & 340$\pm$88 & 24$\pm$17 \\
Non-agreement scorer & 0.875$\pm$0.010 & 0.953$\pm$0.014 & 0.373$\pm$0.144 & 870$\pm$349 & 42$\pm$25 \\
BioConCal logistic & 0.895$\pm$0.010 & 0.948$\pm$0.017 & 0.471$\pm$0.049 & 1096$\pm$116 & 59$\pm$24 \\
\textbf{BioConCal GBT} & \textbf{0.907$\pm$0.010} & \textbf{0.944$\pm$0.013} & \textbf{0.565$\pm$0.064} & \textbf{1321$\pm$161} & \textbf{75$\pm$25} \\
\bottomrule
\end{tabular}
\caption{Robustness over 10 document-level splits (seeds 13, 21, 42, 87, 101, 7, 123, 256, 512, 1000), mean$\pm$sd. Same protocol, gold-free features, and validation-targeted 0.95 precision as Table~\ref{tab:strongcal}; each seed's recall denominators are recomputed from its own test fold. Source: \texttt{scripts/repeated\_docsplit\_bioconcal.py}, \texttt{tables/repeated\_docsplit\_summary.csv}.}
\label{tab:robust}
\end{table*}

\begin{table*}[!t]
\centering
\footnotesize
\begin{tabular}{rcccccc}
\toprule
Seed & AUROC & P@0.95 & R$_{\mathrm{cand}}$ & R$_{\mathrm{corpus}}$ & Sel. & FP \\
\midrule
7 & 0.915 & 0.935 & 0.635 & 0.579 & 1470 & 96 \\
13 & 0.926 & 0.973 & 0.521 & 0.479 & 1134 & 31 \\
21 & 0.907 & 0.948 & 0.508 & 0.459 & 1244 & 65 \\
42 & 0.910 & 0.939 & 0.592 & 0.523 & 1340 & 82 \\
87 & 0.914 & 0.938 & 0.647 & 0.586 & 1514 & 94 \\
101 & 0.903 & 0.953 & 0.519 & 0.464 & 1204 & 57 \\
123 & 0.906 & 0.953 & 0.475 & 0.438 & 1158 & 55 \\
256 & 0.900 & 0.928 & 0.654 & 0.592 & 1578 & 113 \\
512 & 0.890 & 0.944 & 0.528 & 0.475 & 1177 & 66 \\
1000 & 0.900 & 0.932 & 0.574 & 0.515 & 1392 & 94 \\
\midrule
mean$\pm$sd & 0.907$\pm$0.010 & 0.944$\pm$0.013 & 0.565$\pm$0.064 & 0.511$\pm$0.057 & 1321$\pm$161 & 75$\pm$25 \\
\bottomrule
\end{tabular}
\caption{Per-seed BioConCal-GBT results over the 10 document-level splits. Source: \texttt{scripts/repeated\_docsplit\_bioconcal.py}, \texttt{tables/repeated\_docsplit\_all\_seeds.csv}.}
\label{tab:robust_seedwise}
\end{table*}

\begin{table*}[!t]
\centering
\footnotesize
\begin{tabular}{lccc}
\toprule
Scorer & AUROC [95\% CI] & P@0.95 [95\% CI] & R$_{\mathrm{cand}}$ [95\% CI] \\
\midrule
Raw agreement count & 0.753 [0.732, 0.776] & 0.952 [0.928, 0.974] & 0.131 [0.112, 0.151] \\
Vote-vector logistic & 0.864 [0.844, 0.884] & 0.941 [0.914, 0.965] & 0.151 [0.131, 0.174] \\
Aggregate-agreement scorer & 0.861 [0.840, 0.880] & 0.952 [0.928, 0.973] & 0.131 [0.111, 0.153] \\
Non-agreement scorer & 0.865 [0.844, 0.885] & 0.952 [0.849, 1.000] & 0.028 [0.014, 0.044] \\
BioConCal logistic & 0.891 [0.871, 0.910] & 0.931 [0.895, 0.961] & 0.539 [0.499, 0.578] \\
\textbf{BioConCal GBT} & \textbf{0.910 [0.892, 0.928]} & 0.939 [0.908, 0.966] & \textbf{0.592 [0.559, 0.626]} \\
\bottomrule
\end{tabular}
\caption{Document-level bootstrap 95\% confidence intervals on the seed-42 test fold (2{,}000 resamples). The BioConCal-GBT AUROC interval does not overlap the raw-agreement interval. Source: \texttt{scripts/bootstrap\_doclevel\_ci.py}, \texttt{tables/bootstrap\_doclevel\_ci.csv}.}
\label{tab:bootstrap_ci}
\end{table*}

\section{Candidate keying ablation}
\label{app:keying}
\begin{table*}[!t]
\centering
\footnotesize
\begin{tabular}{lrrrrrrrr}
\toprule
Correctness key & Cand. & Prev. & Ceiling & AUROC & P@95 & R$_{\mathrm{cand}}$ & R$_{\mathrm{corpus}}$ & Sel. (FP) \\
\midrule
Normalized text + type & 3{,}866 & 0.549 & 0.883 & 0.910 & 0.939 & 0.592 & 0.523 & 1{,}340 (82) \\
Exact surface + type & 3{,}866 & 0.548 & 0.882 & 0.912 & 0.937 & 0.605 & 0.534 & 1{,}369 (86) \\
Exact span + type & 3{,}866 & 0.098 & 0.158 & 0.858 & 1.000 & 0.032 & 0.005 & 12 (0) \\
\bottomrule
\end{tabular}
\caption{Candidate keying ablation on the document-level 60/20/20 test fold. The correctness key that labels a candidate as positive is varied while the candidate set and gold-free features are held fixed, and BioConCal-GBT is retrained for each key with a validation-selected 0.95 precision threshold. The occurrence-aware normalized row reproduces the canonical Full-8 result. Exact-surface keying is nearly identical, while exact-span keying collapses the positive set and the ceiling because LLM-emitted offsets are unreliable, which is why exact localization is handled by deterministic alignment rather than as a scoring target. Source: \texttt{tables/candidate\_keying\_ablation.csv}.}
\label{tab:keying_app}
\end{table*}

\clearpage
\subsection{Candidate keying / granularity ablation}
\label{app:keying_granularity}
The candidate master is occurrence-level: 18{,}812 rows over 15{,}611 distinct \texttt{(doc, type, normalized\_text)} keys on the four main datasets. The table is a multiset, so repeated mentions and surface variants are kept as distinct rows and not silently merged. If a document mentions \emph{misoprostol} four times and the panel emits the surfaces \emph{misoprostol} and \emph{Misoprostol}, these become separate candidate rows that share the key \texttt{(misoprostol, chemical)} but differ in surface form and occurrence index; each carries its own surface features while agreement is computed per occurrence. Boundary and granularity variants such as \emph{IL-2} versus \emph{IL-2 receptor} normalize to different keys and remain distinct candidates. Across the four main datasets, roughly one row in six is a repeated-mention or surface-variant occurrence rather than a new normalized identity.

Table~\ref{tab:keying_granularity_app} varies the candidate key on the document-level test fold (seed 42) and re-trains BioConCal-GBT. Variants A and B collapse to one row per normalized key and per surface key respectively (representative occurrence with the highest agreement count; gold-positive if any occurrence is); C is the occurrence-aware table actually used; D scores the same rows against the relaxed-overlap label. Coverage is occurrence-level against the same corpus gold ($G_{\mathrm{corpus}}=2{,}404$). Variant~C reproduces the canonical Full-8 numbers (AUROC 0.910, selected 1{,}340), a no-fabrication check. AUROC and the lift over raw agreement are stable across keys, so the triage conclusion does not depend on the keying scheme; what changes is the occurrence-level ceiling (collapsing repeated mentions lowers recoverable gold occurrences from 0.856 to 0.470), a reporting-granularity effect, not a scorer failure. Source: \texttt{scripts/candidate\_keying\_ablation.py}, \texttt{tables/candidate\_keying\_ablation.csv}.

\begin{table}[!t]
\centering
\scriptsize
\resizebox{\columnwidth}{!}{%
\begin{tabular}{lrrcccc}
\toprule
Candidate key & Cands & Prev. & Ceiling & AUROC & Raw AUROC & P@0.95 \\
\midrule
A. normalized-collapsed & 12123 & 0.471 & 0.470 & 0.908 & 0.836 & 0.958 \\
B. surface key & 12580 & 0.483 & 0.505 & 0.911 & 0.833 & 0.945 \\
C. occurrence-aware (current) & 18812 & 0.593 & 0.856 & 0.910 & 0.753 & 0.939 \\
D. relaxed-overlap label & 18812 & 0.758 & 0.856 & 0.862 & 0.745 & 0.943 \\
\bottomrule
\end{tabular}
}
\vspace{-6pt}
\caption{Candidate keying / granularity ablation. Variant C is the occurrence-aware canonical setting.}
\label{tab:keying_granularity_app}
\end{table}

\section{Full LODO failure analysis}
\label{app:lodo_fail}
\begin{table*}[!t]
\centering
\scriptsize
\begin{tabular}{lp{0.15\textwidth}p{0.17\textwidth}rrrrrr}
\toprule
Held-out & Types in test & Unseen-in-train & Prev. & Raw & GBT & Log. & P@95 & R$_{\mathrm{cand}}$@95 \\
& & & & AUROC & AUROC & AUROC & test & test \\
\midrule
BC5CDR & chem., disease & chem. & 0.653 & 0.737 & 0.781 & 0.846 & 1.000 & 0.001 \\
NCBI Disease & disease & --- & 0.506 & 0.869 & 0.910 & 0.919 & 0.979 & 0.100 \\
BC2GM & gene & gene & 0.406 & 0.766 & 0.857 & 0.828 & 0.771 & 0.415 \\
JNLPBA & protein, DNA, RNA, cell line, cell type & DNA, RNA, cell line, cell type & 0.399 & 0.794 & 0.853 & 0.837 & 0.889 & 0.165 \\
\bottomrule
\end{tabular}
\caption{Full LODO per-fold table, separating ranking transfer (raw and BioConCal AUROC) from operating-point transfer (P@95 and R$_{\mathrm{cand}}$@95 from a source-selected threshold). ``Unseen-in-train'' lists entity types in the held-out fold but not in training. Sources: \texttt{tables/lodo\_ranking\_vs\_threshold\_transfer.csv}, \texttt{tables/lodo\_failure\_analysis.csv}.}
\label{tab:lodo_fail_app}
\end{table*}

\paragraph{Deployment checklist.}
\label{app:deploy}
Before applying BioConCal to a new corpus we recommend the following steps. (1) Verify that the target entity types are represented in the validation data. (2) Estimate the target-domain candidate positive prevalence. (3) Select operating thresholds on target-domain validation documents. (4) Report empirical precision on a held-out target audit set before use. (5) Use scores for triage only, not for automatic database insertion. (6) If target entity types are unseen in training or calibration, treat a source-selected 0.95-precision threshold as invalid.

\section{LODO target-domain recalibration}
\label{app:lodo_recal}
\begin{table*}[!t]
\centering
\footnotesize
\begin{tabular}{lrrrrr}
\toprule
Held-out & $n_{cal}$ & Test P@95 & Test R$_{\mathrm{cand}}$@95 & Sel. & Notes \\
\midrule
BC5CDR & 0 & 1.000 & 0.001 & 5 & source-val (canonical v3 LODO) \\
BC5CDR & 25 & 0.859 $\pm$ 0.100 & 0.004 $\pm$ 0.003 & 45 & target-cal, 3 seeds \\
BC5CDR & 50 & 0.828 $\pm$ 0.063 & 0.005 $\pm$ 0.003 & 46 & target-cal, 3 seeds \\
BC5CDR & 100 & 0.862 $\pm$ 0.090 & 0.003 $\pm$ 0.002 & 30 & target-cal, 3 seeds \\
NCBI Disease & 0 & 0.979 & 0.100 & 47 & source-val (canonical v3 LODO) \\
NCBI Disease & 25 & 0.924 $\pm$ 0.007 & 0.682 $\pm$ 0.033 & 319 & target-cal, 3 seeds \\
NCBI Disease & 50 & 0.904 $\pm$ 0.030 & 0.719 $\pm$ 0.086 & 323 & target-cal, 3 seeds \\
NCBI Disease & 100 & 0.890 $\pm$ 0.037 & 0.737 $\pm$ 0.105 & 291 & target-cal, 3 seeds \\
BC2GM & 0 & 0.771 & 0.415 & 280 & source-val (canonical v3 LODO) \\
BC2GM & 25 & 0.830 $\pm$ 0.023 & 0.079 $\pm$ 0.080 & 47 & target-cal, 3 seeds \\
BC2GM & 50 & 0.555 $\pm$ 0.393 & 0.019 $\pm$ 0.014 & 10 & target-cal, 3 seeds \\
BC2GM & 100 & 0.544 $\pm$ 0.384 & 0.019 $\pm$ 0.014 & 9 & target-cal, 3 seeds \\
JNLPBA & 0 & 0.889 & 0.165 & 199 & source-val (canonical v3 LODO) \\
JNLPBA & 25 & 0.875 $\pm$ 0.008 & 0.296 $\pm$ 0.099 & 345 & target-cal, 3 seeds \\
JNLPBA & 50 & 0.867 $\pm$ 0.009 & 0.240 $\pm$ 0.132 & 268 & target-cal, 3 seeds \\
JNLPBA & 100 & 0.903 $\pm$ 0.028 & 0.066 $\pm$ 0.045 & 64 & target-cal, 3 seeds \\
\bottomrule
\end{tabular}
\caption{LODO with target-domain recalibration. The $n_{cal}{=}0$ rows are canonical source-validation LODO results; $n_{cal}\geq25$ rows reuse the same source-trained model, select the P95 threshold on target-domain calibration documents, and evaluate on the remaining target documents. Mean$\pm$sd over three seeds. Source: \texttt{tables/lodo\_target\_recalibration\_summary.csv}.}
\label{tab:lodo_recal_app}
\end{table*}

\section{Gain decomposition}
\label{app:gain_decomp}
\begin{center}
\begin{minipage}{\columnwidth}
\centering
\footnotesize
\begin{tabular}{llrrr}
\toprule
Panel & Features & AUROC & P@95 & R$_{full}$@95 \\
\midrule
Best open & NA & 0.880 & 0.933 & 0.262 \\
Closed-2 & NA & 0.794 & 0.955 & 0.160 \\
Closed-2 & NA+AG & 0.806 & 0.943 & 0.286 \\
Open-6 & NA & 0.899 & 0.944 & 0.331 \\
Open-6 & NA+AG & 0.908 & 0.958 & 0.335 \\
Full-8 & NA & 0.865 & 0.952 & 0.028 \\
Full-8 & AG & 0.873 & 0.951 & 0.387 \\
Full-8 & NA+AG & 0.910 & 0.939 & 0.592 \\
\bottomrule
\end{tabular}
\captionof{table}{Gain decomposition on the document-level 60/20/20 test fold. NA is the non-agreement feature block; AG is the agreement-structure block. Source: \texttt{tables/gain\_decomposition.csv}.}
\label{tab:gain_decomp_app}
\end{minipage}
\end{center}

\clearpage
\section{Full cheap-panel baseline table}
\label{app:cheap_panel}
\begin{table*}[!t]
\centering
\footnotesize
\begin{tabular}{lrrrrrrrrrrrr}
\toprule
Panel & $|M|$ & Cand. & Prev. & AUROC & AUPRC & Brier & ECE & Sel. & FP & Test P & Cand. R & Cand. Cov \\
\midrule
OpenAI & 1 & 12{,}628 & 0.793 & 0.815 & 0.937 & 0.134 & 0.057 & 368 & 6 & 0.984 & 0.177 & 0.141 \\
Claude & 1 & 9{,}324 & 0.831 & 0.814 & 0.945 & 0.116 & 0.065 & 604 & 18 & 0.970 & 0.373 & 0.317 \\
Best open & 1 & 7{,}979 & 0.674 & 0.880 & 0.916 & 0.134 & 0.062 & 596 & 40 & 0.933 & 0.516 & 0.352 \\
Closed-2 & 2 & 13{,}698 & 0.769 & 0.806 & 0.915 & 0.140 & 0.063 & 644 & 37 & 0.943 & 0.278 & 0.224 \\
Diverse-3 & 3 & 15{,}156 & 0.687 & 0.880 & 0.926 & 0.131 & 0.046 & 1{,}003 & 56 & 0.944 & 0.490 & 0.338 \\
Open-6 & 6 & 14{,}474 & 0.564 & 0.908 & 0.914 & 0.125 & 0.055 & 742 & 31 & 0.958 & 0.458 & 0.247 \\
Full-8 & 8 & 18{,}812 & 0.593 & 0.910 & 0.923 & 0.119 & 0.032 & 1{,}340 & 82 & 0.939 & 0.592 & 0.347 \\
\bottomrule
\end{tabular}
\caption{Full cheap-panel baseline results on the document-level 60/20/20 test fold. Prev. is positive prevalence in the panel candidate universe. Sel., FP, and Test P are at the validation-selected P95 operating point. Test P is empirical test precision. Cand. R and Cand. Cov are candidate recall and candidate coverage over the panel's own universe. The main-text Table~\ref{tab:cheap_panel} reports R$_{full}$@95, which is recall over the full eight-panel positive candidate set. Source: \texttt{tables/single\_model\_candidate\_baselines\_p95.csv}.}
\label{tab:cheap_panel_app}
\end{table*}

\section{End-to-end interpretation}
\label{app:e2e}
\begin{table}[!htbp]
\centering
\scriptsize
\begin{tabular}{lrrrrr}
\toprule
Method & Sel. & FP & C-P & C-R & Corp-R \\
\midrule
Union (any model) & 3{,}866 & 1{,}742 & 0.549 & 1.000 & 0.883 \\
OpenAI $\cup$ Claude & 2{,}743 & 743 & 0.729 & 0.942 & 0.832 \\
Agree $\geq$ 2 & 2{,}587 & 797 & 0.692 & 0.843 & 0.745 \\
Agree $\geq$ 4 & 1{,}275 & 260 & 0.796 & 0.478 & 0.422 \\
Agree $\geq$ 6 & 715 & 93 & 0.870 & 0.293 & 0.259 \\
Unanimous ($k{=}8$) & 293 & 14 & 0.952 & 0.131 & 0.116 \\
\textbf{BioConCal P95} & \textbf{1{,}340} & \textbf{82} & \textbf{0.939} & \textbf{0.592} & \textbf{0.523} \\
\bottomrule
\end{tabular}
\caption{End-to-end interpretation on the doc-level 60/20/20 test fold. C-P is candidate precision. C-R is candidate recall (selected TP over 2{,}124 gold-positive candidates). Corp-R is corpus-level recall (selected TP over 2{,}404 gold mentions in the test corpus). The empirical union candidate recall is 0.883, which caps Corp-R for any candidate-level method. Source: \texttt{tables/end\_to\_end\_interpretation.csv}.}
\label{tab:e2e_app}
\end{table}

\begin{table*}[!t]
\centering
\scriptsize
\begin{tabular}{p{0.17\textwidth}p{0.12\textwidth}rrrrrcp{0.23\textwidth}}
\toprule
Workflow & Cand.\ source & Test P & TP & FP & Sel. & R$_{\mathrm{corpus}}$ & Triage fit & Main interpretation \\
\midrule
Unanimous / raw agreement P95 & LLM-8 & 0.952 & 279 & 14 & 293 & 0.116 & Low yield & High precision, but too conservative for review yield \\
Agreement $\geq$ 2 & LLM-8 & 0.692 & 1{,}790 & 797 & 2{,}587 & 0.745 & Low precision & High raw recall, but 797 FPs make the queue noisy \\
BERN2 take-all & BERN2 & 0.806 & 1{,}772 & 427 & 2{,}199 & 0.737 & Unfiltered & Higher coverage, but no learned candidate filter \\
LLM-8 $\cup$ BERN2 exploratory merge & additive union & 0.836 & 1{,}323 & 260 & 1{,}583 & 0.550 & Unfiltered & Raises ceiling, but unfiltered additions add FPs \\
\textbf{BioConCal GBT} & LLM-8 & \textbf{0.939} & \textbf{1{,}258} & 82 & 1{,}340 & 0.522 & \textbf{Primary} & Largest primary high precision review yield \\
BioConCal-Hybrid & merged union & 0.929 & 1{,}716 & 131 & 1{,}847 & 0.709 & Tradeoff & Higher coverage, but lower precision than LLM-8 BioConCal \\
\bottomrule
\end{tabular}
\caption{Curator triage utility under high precision review constraints on the document level test fold. Rows compare fixed voting, unfiltered source expansion, and learned scoring workflows by empirical precision and selected yield. The primary BioConCal operating point is selected on validation at the 0.95 precision target. TP denotes selected candidate row positives. Hybrid R$_{\mathrm{corpus}}$ uses multiset corpus accounting, with 1{,}705 selected gold occurrences over 2{,}404 corpus gold mentions. BioConCal GBT is the primary LLM-8 review queue. BioConCal-Hybrid scores the expanded LLM-8\,$\cup$\,BERN2 universe and trades precision for coverage. This is a curator triage comparison, not an end-to-end extraction. Data sources are \texttt{tables/risk\_control\_selection\_docsplit.csv}, \texttt{tables/end\_to\_end\_interpretation.csv}, \texttt{tables/main\_candidate\_generator\_comparison.csv}, and \texttt{tables/hybrid\_curator\_utility.csv}.}
\label{tab:failmodes}
\end{table*}

\section{Reliability diagram}
\label{app:reliability}
\begin{figure}[!htbp]
\centering
\includegraphics[width=0.9\linewidth]{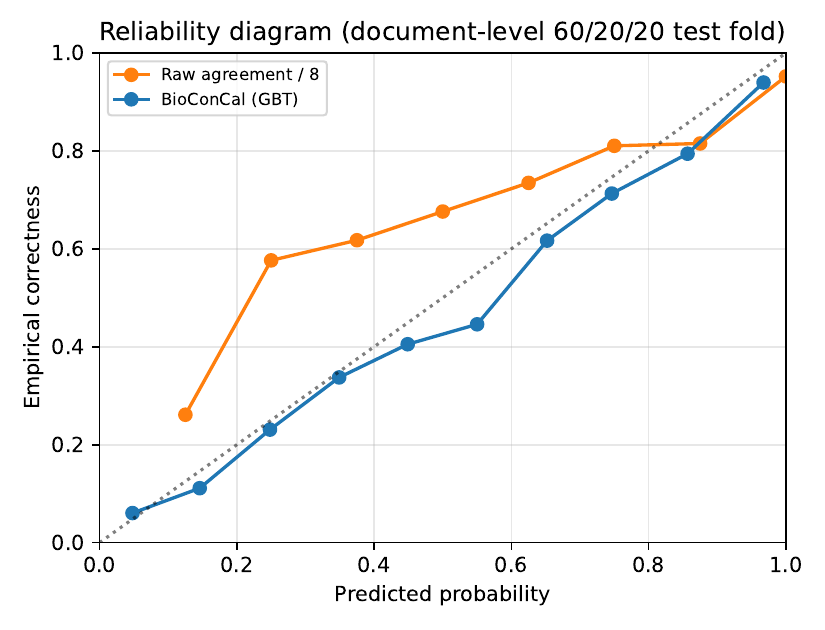}
\caption{Reliability diagram on the document-level 60/20/20 test fold.}
\label{fig:reliability_app}
\end{figure}

\section{Agreement-count calibration curve}
\label{app:agreement}
\begin{figure}[!htbp]
\centering
\includegraphics[width=\linewidth]{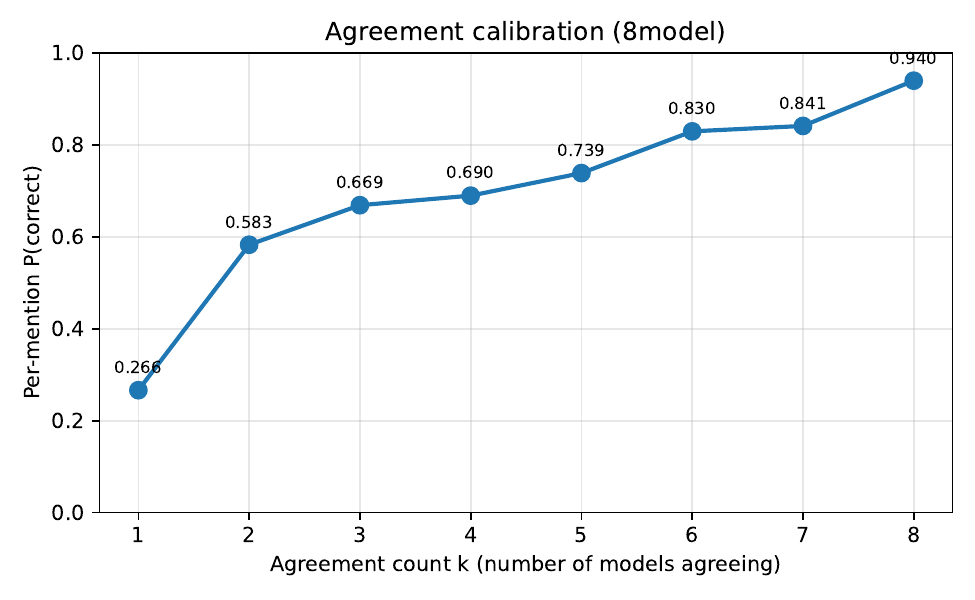}
\caption{Per-mention $P(\text{correct} \mid k)$ as a function of agreement count $k$ on the 8-model panel. Precision rises monotonically from 0.266 at $k{=}1$ to 0.940 at $k{=}8$.}
\label{fig:agreement}
\end{figure}

\section{Full selective and conformal baselines}
\label{app:selective}
\begin{table*}[!t]
\centering
\footnotesize
\begin{tabular}{lrrrrr}
\toprule
Method & Sel. & FP & Test P & Test R & P gap \\
\midrule
Raw agreement (val P95) & 293 & 14 & 0.952 & 0.131 & $+0.002$ \\
Conformal-style raw cutoff ($\alpha{=}0.05$) & 715 & 93 & 0.870 & 0.293 & $-0.080$ \\
Platt scaling on $k$ & 293 & 14 & 0.952 & 0.131 & $+0.002$ \\
Isotonic on $k$ & 293 & 14 & 0.952 & 0.131 & $+0.002$ \\
BioConCal GBT (val P95) & 1{,}340 & 82 & 0.939 & 0.592 & $-0.011$ \\
Conformal-style BioConCal cutoff ($\alpha{=}0.05$) & 1{,}362 & 83 & 0.939 & 0.602 & $-0.011$ \\
\bottomrule
\end{tabular}
\caption{Selective and split-conformal-inspired baselines on the document-level 60/20/20 split. ``P gap'' is the empirical test precision minus the 0.95 nominal target. Source: \texttt{tables/selective\_baselines.csv}. These are comparison baselines, not formal finite-sample precision guarantees.}
\label{tab:selective_app}
\end{table*}

\section{Panel composition ablation}
\label{app:panel}
\begin{table*}[!t]
\centering
\footnotesize
\begin{tabular}{lrrrrrr}
\toprule
Panel & N$_\text{cand}$ & AUROC & Brier & ECE & P@95 & R$_{\mathrm{cand}}$@95 \\
\midrule
All 8 models & 18{,}812 & 0.910 & 0.119 & 0.032 & 0.939 & 0.592 \\
Open-weight 6 & 14{,}474 & 0.908 & 0.142 & 0.069 & 0.958 & 0.458 \\
Closed 2 (OpenAI+Claude) & 13{,}698 & 0.806 & 0.211 & 0.117 & 0.943 & 0.278 \\
Drop OpenAI & 16{,}977 & 0.911 & 0.121 & 0.038 & 0.937 & 0.632 \\
Drop Claude & 18{,}174 & 0.909 & 0.124 & 0.039 & 0.933 & 0.574 \\
\bottomrule
\end{tabular}
\caption{Panel-composition ablation of BioConCal-GBT under the document-level 60/20/20 split. Agreement features are recomputed from each panel subset; candidates with zero in-panel votes are removed. The closed-only two-model panel drops AUROC to 0.806, consistent with the importance of model diversity. The full leave-one-model-out grid is stable in the 0.899--0.912 AUROC range and is in \texttt{tables/panel\_composition\_ablation.csv}.}
\label{tab:panel_app}
\end{table*}

\section{Unanimous false-positive audit (auto-suggested)}
\label{app:fp_audit}
\begin{center}
\begin{minipage}{\columnwidth}
\centering
\footnotesize
\begin{tabular}{lrr}
\toprule
Auto-suggested category & Count & \% \\
\midrule
Confirmed false positive & 0 & 0.0 \\
Boundary mismatch & 66 & 73.3 \\
Type confusion & 7 & 7.8 \\
Alias / synonym not in gold & 13 & 14.4 \\
Broader / narrower concept & 0 & 0.0 \\
Possible gold omission & 3 & 3.3 \\
Multiset overflow & 1 & 1.1 \\
\midrule
Total & 90 & 100.0 \\
\bottomrule
\end{tabular}
\captionof{table}{Auto-suggested category distribution for the 90 unanimous (8-model) false positives. Categories are assigned by a rule-based heuristic and have not been manually verified. ``Multiset overflow'' is the case where the panel emitted the same $(\text{normalized text}, \text{type})$ key more times than gold contained. We also release a 50-case stratified manual-audit sample (\texttt{audit/unanimous\_fp\_manual\_sample\_50.csv}) with blank \texttt{manual\_category} and \texttt{manual\_notes} columns.}
\label{tab:fp_audit_app}
\end{minipage}
\end{center}

\section{Feature robustness ablation}
\label{app:featrob}
\begin{table*}[!t]
\centering
\footnotesize
\begin{tabular}{lrrrrrrr}
\toprule
Variant (GBT) & \#feat. & AUROC & $\Delta$AUROC & Brier & P@95 & R$_{\mathrm{cand}}$@95 & Sel \\
\midrule
A. BioConCal main model & 19 & 0.910 & $+0.000$ & 0.119 & 0.939 & 0.592 & 1{,}340 \\
B. remove entity\_type & 18 & 0.906 & $-0.004$ & 0.122 & 0.940 & 0.559 & 1{,}264 \\
C. remove doc\_text\_length & 18 & 0.909 & $-0.001$ & 0.119 & 0.942 & 0.549 & 1{,}238 \\
D. remove surface availability & 15 & 0.910 & $+0.000$ & 0.120 & 0.935 & 0.599 & 1{,}361 \\
E. remove offset only & 17 & 0.910 & $+0.000$ & 0.120 & 0.935 & 0.599 & 1{,}361 \\
F. remove doc length + surface avail. & 14 & 0.912 & $+0.002$ & 0.117 & 0.934 & 0.600 & 1{,}364 \\
G. agreement + mention only & 13 & 0.909 & $-0.001$ & 0.120 & 0.941 & 0.528 & 1{,}193 \\
H. agreement only & 4 & 0.860 & $-0.050$ & 0.149 & --- & 0.000 & 0 \\
I. vote-pattern only & 8 & 0.859 & $-0.051$ & 0.151 & 0.953 & 0.143 & 318 \\
\bottomrule
\end{tabular}
\caption{Feature-removal robustness ablation under the doc-level 60/20/20 split and val-frozen P95 protocol. $\Delta$AUROC is relative to variant A, the main 19-feature BioConCal model. AUROC and R$_{\mathrm{cand}}$@95 are stable across A--G (max $|\Delta\text{AUROC}|=0.004$); only stripping to agreement-only (H) or vote-pattern-only (I) collapses precision-targeted selection. The non-agreement features (mention/surface/document) are useful in-domain but not the dominant signal, and the scorer is not driven by corpus-specific formatting shortcuts. Source: \texttt{tables/feature\_robustness\_ablation.csv}.}
\label{tab:featrob_app}
\end{table*}

\section{Curator workload trade-off}
\label{app:curator}
\begin{table*}[!t]
\centering
\footnotesize
\begin{tabular}{lrrrrrrr}
\toprule
Method & Sel. & TP & FP & $\Delta$TP & $\Delta$FP & $\Delta$FP/$\Delta$TP & Reviewed/$\Delta$TP \\
\midrule
Unanimous ($k{=}8$) & 293 & 279 & 14 & 0 & 0 & 0 & 0 \\
BioConCal GBT (val P95) & 1{,}340 & 1{,}258 & 82 & 979 & 68 & 0.069 & 1.069 \\
\bottomrule
\end{tabular}
\caption{Curator workload trade-off between Unanimous and BioConCal P95 on the doc-level test fold. ``$\Delta$TP'' and ``$\Delta$FP'' are gains over Unanimous; ``$\Delta$FP/$\Delta$TP'' is the marginal false-positive cost per additional true candidate; ``Reviewed/$\Delta$TP'' is the curator-side review burden per recovered true positive. Source: \texttt{tables/curator\_workload\_tradeoff.csv}.}
\label{tab:curator_app}
\end{table*}

\section{Learned-selector baselines}
\label{app:learned}
\begin{table*}[!t]
\centering
\footnotesize
\begin{tabular}{lrrrrrr}
\toprule
Selector & AUROC & Brier & P@95 & R$_{\mathrm{cand}}$@95 & Sel & FP \\
\midrule
Raw agreement threshold & 0.753 & 0.237 & 0.952 & 0.131 & 293 & 14 \\
Agreement-count logistic & 0.753 & 0.205 & 0.952 & 0.131 & 293 & 14 \\
Agreement-count GBT & 0.751 & 0.199 & 0.952 & 0.131 & 293 & 14 \\
Vote-pattern logistic & 0.864 & 0.148 & 0.941 & 0.151 & 341 & 20 \\
Vote-pattern GBT & 0.859 & 0.151 & 0.953 & 0.143 & 318 & 15 \\
Agreement-family GBT & 0.860 & 0.149 & --- & --- & 0 & 0 \\
Agreement + mention GBT & 0.909 & 0.120 & 0.941 & 0.528 & 1{,}193 & 71 \\
\textbf{BioConCal GBT (main 19-feature)} & \textbf{0.910} & \textbf{0.119} & \textbf{0.939} & \textbf{0.592} & \textbf{1{,}340} & \textbf{82} \\
\bottomrule
\end{tabular}
\caption{Learned-selector baselines under the same doc-level 60/20/20 split and val-targeted P95 protocol. Vote-pattern (8 per-model indicators) and aggregate-agreement (4-feature) GBT classifiers both reach AUROC $\approx 0.86$ but recover little precision-targeted recall; the largest R$_{\mathrm{cand}}$@95 jump comes from adding mention features. The agreement-family GBT row selects zero candidates because no val-fold operating point of that classifier reaches the 0.95 precision target. Source: \texttt{tables/learned\_selectors\_baseline.csv}.}
\label{tab:learned_app}
\end{table*}

\begin{figure*}[!t]
\centering
\IfFileExists{figures/figure_precision_coverage_docsplit.pdf}{\includegraphics[width=0.7\linewidth]{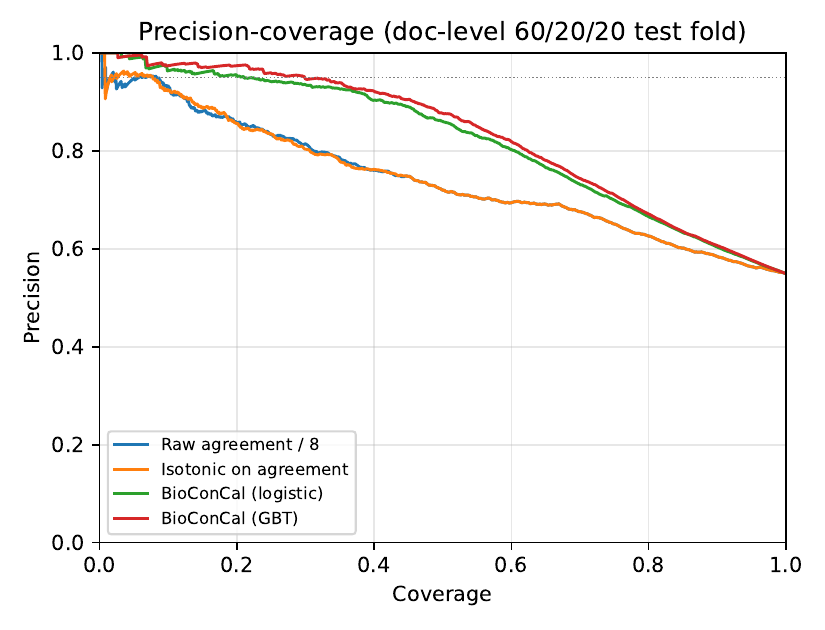}}{\fbox{\parbox{0.90\linewidth}{Precision-coverage curve missing.}}}
\caption{Precision-coverage curves on the document-level 60/20/20 test fold. BioConCal improves on raw agreement count across the coverage range.}
\label{fig:prcov}
\end{figure*}

\section{Span alignment per selection rule}
\label{app:span}
\begin{center}
\begin{minipage}{\columnwidth}
\centering
\footnotesize
\begin{tabular}{lrrrr}
\toprule
Selection rule & Stage & P & R & F1 \\
\midrule
OpenAI alone & before & 0.075 & 0.077 & 0.076 \\
OpenAI alone & \textbf{after} & 0.765 & 0.789 & \textbf{0.777} \\
Claude alone & before & 0.188 & 0.143 & 0.162 \\
Claude alone & \textbf{after} & 0.819 & 0.623 & \textbf{0.708} \\
MiniMax-M2 alone & before & 0.019 & 0.013 & 0.015 \\
MiniMax-M2 alone & \textbf{after} & 0.509 & 0.332 & \textbf{0.402} \\
Unanimous ($k=8$) & after & 0.924 & 0.113 & 0.201 \\
BioConCal @ P=0.95 & after & 0.666 & 0.175 & 0.277 \\
\bottomrule
\end{tabular}
\captionof{table}{Exact-span F1 before and after deterministic span realignment. Source: \texttt{tables/span\_alignment\_results.csv}.}
\label{tab:span}
\end{minipage}
\end{center}

\section{Open-weight prompt sensitivity}
\label{app:promptsens}
We evaluated prompt sensitivity on a fixed 50-document-per-dataset subset (seed 42) of the four main datasets. The complete three-model open-weight panel used for the cross-prompt agreement-calibration analysis is Phi-4 14B, Mistral-Small 3.2 24B, and MiniMax-M2 230B; these three models ran all three prompts to completion. Llama 3.3 70B (served via Ollama) finished only the original prompt and is therefore partial; it contributes to the single-model table for the original prompt but is excluded from the cross-prompt agreement calibration to keep the panel size constant across prompts. CHEMDNER is excluded from this analysis because the main paper already treats it as robustness-only (its public mirror does not provide reliable character offsets). These results are not claimed as closed-model robustness or as full eight-model prompt sensitivity; closed-API models (OpenAI GPT-5.5 and Claude Opus 4.7) were not re-run with prompt variants. Three prompts are compared: \emph{original\_zero\_shot} (the main paper's template), \emph{strict\_entity\_only} (instructs the model to extract only explicit surface mentions and not infer entities), and \emph{few\_shot\_2\_examples} (two synthetic illustrative examples that do not overlap the evaluation subset).

\begin{table}[!htbp]
\centering
\scriptsize
\setlength{\tabcolsep}{1pt}
\begin{tabular}{@{}lrrrrr@{}}
\toprule
Prompt & Cands & P@1 & P@2 & P@max & Agr. AUC \\
\midrule
original\_zero\_shot & 797 & 0.251 & 0.585 & 0.876 & 0.778 \\
strict\_entity\_only & 649 & 0.392 & 0.704 & 0.919 & 0.742 \\
few\_shot\_2\_examples & 913 & 0.241 & 0.589 & 0.776 & 0.753 \\
\bottomrule
\end{tabular}
\caption{Open-weight prompt sensitivity on the three-model panel (Phi-4, Mistral-Small 3.2, MiniMax-M2; max agreement $k{=}3$). $P(\text{correct}\mid k)$ is monotonic in $k$ for all three prompts. Strict prompting raises high-agreement precision at the cost of recall; few-shot prompting raises recall at the cost of P@max~$k$. Source: \texttt{tables/prompt\_sensitivity\_summary.csv}.}
\label{tab:promptsens}
\end{table}

\begin{figure}[!htbp]
\centering
\includegraphics[width=\linewidth]{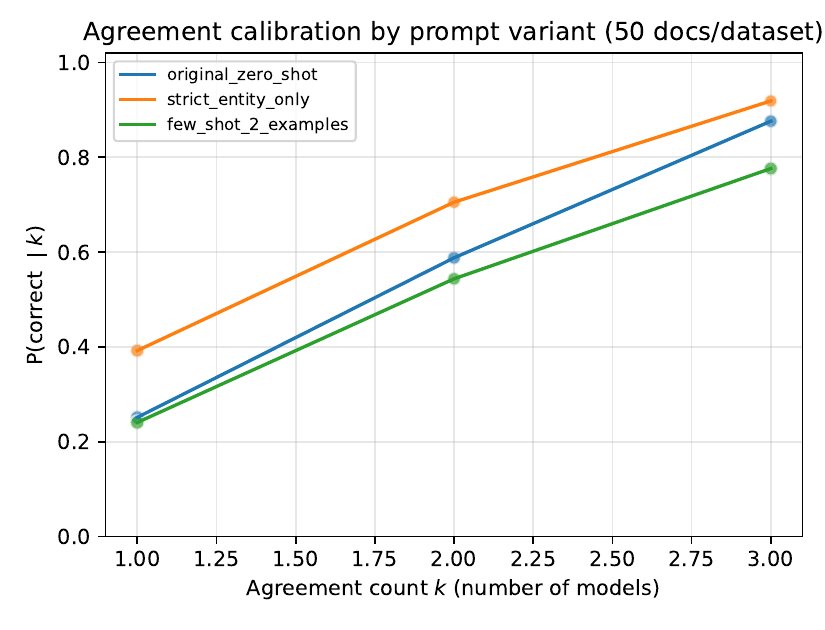}
\caption{Agreement calibration across the three prompt variants on the open-weight 50-doc-per-dataset subset. Marker size scales with the number of candidates at each agreement count.}
\label{fig:promptsens}
\end{figure}

\section{Per-dataset candidate coverage}
\label{app:ceiling}
\begin{table*}[!t]
\centering
\footnotesize
\begin{tabular}{lrrrrrr}
\toprule
Dataset & Gold & Represented & Ceiling & Union cands & BC Sel & BC Corp-R \\
\midrule
BC5CDR & 1{,}939 & 1{,}643 & 0.847 & 3{,}398 & 1{,}228 & 0.594 \\
NCBI Disease & 108 & 95 & 0.880 & 175 & 47 & 0.426 \\
BC2GM & 130 & 105 & 0.808 & 184 & 13 & 0.069 \\
JNLPBA & 227 & 214 & 0.943 & 558 & 52 & 0.225 \\
\midrule
Test total & 2{,}404 & 2{,}057 & 0.856 & 3{,}866 & 1{,}340 & 0.523 \\
\bottomrule
\end{tabular}
\caption{Per-dataset candidate coverage on the doc-level 60/20/20 test fold. ``Gold'' is the total gold mentions in the test docs; ``Represented'' counts gold mentions whose $(\text{normalized text},\text{type})$ key appears in the candidate universe. Two numbers can be reported here: the \emph{stricter multiset gold-key coverage} (0.856 total, where a gold mention only counts as represented if a candidate has the same key under multiset bookkeeping), and the \emph{empirical union candidate recall under candidate-level matching} used in the main text (0.883 total). BC (BioConCal) Corp-R is selected TP / dataset gold mentions. Source: \texttt{tables/candidate\_universe\_ceiling.csv}.}
\label{tab:ceiling_app}
\end{table*}

\section{Per-dataset agreement calibration and normalization}
\label{app:norm_app}
\begin{figure}[!htbp]
\centering
\includegraphics[width=0.95\linewidth]{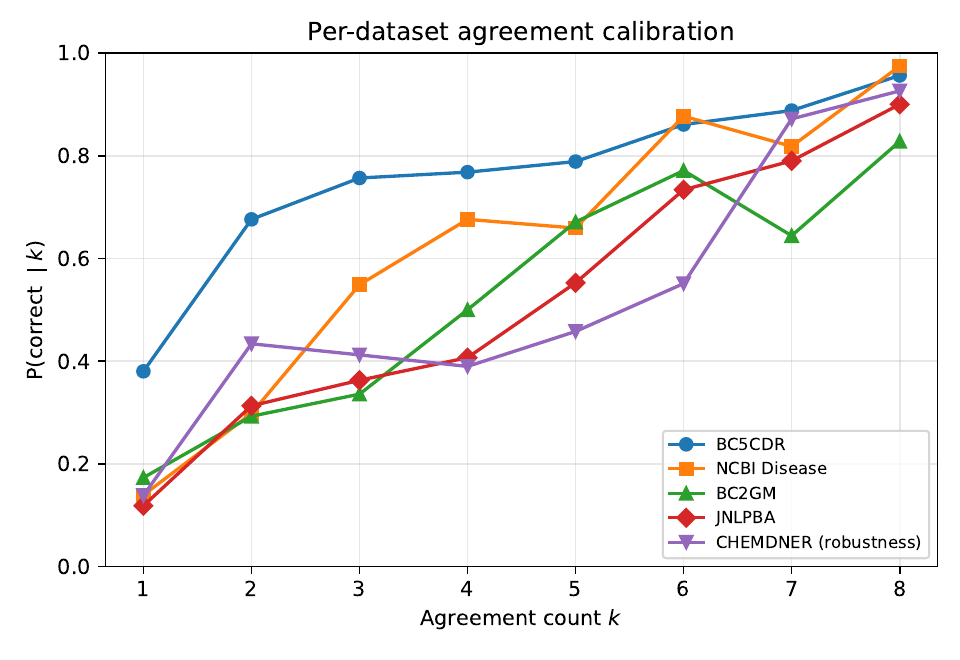}
\caption{Agreement calibration broken down by dataset. BC5CDR and NCBI Disease reach 0.96 and 0.97 at $k{=}8$. BC2GM reaches 0.83, JNLPBA reaches 0.90. Source: \texttt{tables/agreement\_calibration\_by\_type.csv} and \texttt{figures/figure\_calibration\_by\_dataset.pdf}.}
\label{fig:cal_by_ds_app}
\end{figure}

Per-unified-type agreement calibration is released as \texttt{tables/agreement\_calibration\_by\_type.csv}. The normalization ablation comparing exact-surface, normalized, and relaxed-overlap matching is in \texttt{tables/normalization\_ablation.csv}. The mean F1 lift from exact-surface to normalized matching is small across all eight models. The lift from normalized to relaxed-overlap is larger but does not change the model ranking, so the central findings hold under either matching variant.

\end{document}